%% file: main.tex
\documentclass[12pt]{article}
\usepackage[utf8]{inputenc}
\usepackage{amsmath}
\usepackage{amssymb}
\usepackage{amsthm}
\usepackage{amsfonts}
\usepackage{bm}
\usepackage{float}
\usepackage{graphicx}
\usepackage{microtype}
\usepackage{times}
\usepackage[dvipsnames]{xcolor}
\usepackage[hidelinks]{hyperref}
\hypersetup{
    colorlinks=true,
    linkcolor={blue!50!black},
    citecolor={blue!50!black},
    urlcolor={blue!80!black}
}
\usepackage{authblk}

\usepackage[
backend=biber,
style=science
]{biblatex}
\usepackage{algorithm,algorithmicx,algpseudocode}
\algblock{Input}{EndInput}
\algnotext{EndInput}
\algblock{Output}{EndOutput}
\algnotext{EndOutput}
\newcommand{\Desc}[2]{\State \makebox[2em][l]{#1}#2}

\newenvironment{sciabstract}{%
\begin{quote} \bf}
{\end{quote}}

\usepackage{lineno}
\usepackage{etoolbox}
\apptocmd{\thebibliography}{\raggedright}{}{}
\input{math_commands.tex}

\input{macros.tex}
\title{
    The AI Economist: Optimal Economic Policy Design via Two-level Deep Reinforcement Learning
}
\author[*,$\dagger$,1]{Stephan Zheng}
\author[*,1]{Alexander Trott}
\author[1]{Sunil Srinivasa}
\author[1,2]{David C. Parkes}
\author[3]{Richard Socher}
\affil[1]{Salesforce Research}
\affil[2]{Harvard University}
\affil[3]{You.com}
\date{\today}
\begin{document}
\newrefsection
\maketitle
\input{src/v2/abstract.tex}
{}\let\thefootnote\relax\footnotetext{
*: equal contribution.
${}^\dagger$: Correspondence to: \url{stephan.zheng@salesforce.com}.
}
\newpage
{}\let\thefootnote\relax\footnotetext{
A.T. and S.Z. contributed equally. R.S. and S.Z. conceived and directed the project; S.Z., A.T., and D.P. developed the theoretical framework; A.T., S.S., and S.Z. developed the economic simulator, implemented the reinforcement learning platform, and performed experiments; A.T., S.Z., and D.P. processed and analyzed experiments with AI agents; S.Z., A.T., and D.P. drafted the manuscript; R.S. planned and advised the work, and analyzed all results; All authors discussed the results and commented on the manuscript.
}
\input{src/v2/intro.tex}
\input{src/v2/main_body.tex}
\input{src/v2/discussion.tex}
\input{src/v2/ethics.tex}
\input{src/v2/methods.tex}

\Urlmuskip=0mu plus 1mu\relax
\printbibliography[title=References]

\input{src/v2/end-notes.tex}

\end{document}

%% file: math_commands.tex
\def\eqref#1{equation~\ref{#1}}
\def\va{{\bm{a}}}

\DeclareMathAlphabet{\mathsfit}{\encodingdefault}{\sfdefault}{m}{sl}
\SetMathAlphabet{\mathsfit}{bold}{\encodingdefault}{\sfdefault}{bx}{n}

\def\gP{{\mathcal{P}}}

\def\sA{{\mathbb{A}}}

\newcommand{\E}{\mathbb{E}}

\newcommand{\V}[1]{\bm{#1}}

\newcommand{\eq}[1]{\begin{align}#1\end{align}}

\newcommand{\fr}[2]{\frac{#1}{#2}}
\newcommand{\brck}[1]{\left(#1\right)}

\newcommand{\brcksq}[1]{\left[#1\right]}

%% file: macros.tex
\def\ExtendedDataString{}
\def\mweight{\theta}
\def\pweight{\phi}

\def\GatherTradeBuild{Gather-Trade-Build}
\def\height{n_h}
\def\width{n_w}
\def\channels{n_c}

\newcommand{\policy}{\pi}
\def\policyweight{\theta}

\def\ob{o}
\def\Ac{\va}
\def\ac{a}

\def\st{s}

\def\rew{r}

\def\df{\gamma}

\def\eplen{H}
\def\samplinghorizon{\mathcal{H}}
\def\trans{\gP}
\def\hidden{h}

\def\dataset{\mathcal{D}}
\def\stoppingcriterion{\mathcal{C}}
\def\numberofagents{N}

\def\util{u}

\def\endow{x}

\newcommand{\posttax}[1]{\tilde{#1}}

\def\skill{\nu}
\def\elas{e}

\def\income{z}
\def\totalincome{Z}

\def\labor{l}
\def\totallabor{L}

\def\isoeta{\eta}
\def\laborexponent{\delta}
\def\laborcoeff{c}

\def\tax{T}
\def\mtaxrate{\tau}
\def\taxperiodlen{\mathcal{T}}

\def\bracketcutoff{b}
\def\numbrackets{B}

\def\incometax{\tax}

\def\money{C}

\def\vmoney{\V{\money}}

\def\pdf{f}
\def\cdf{F}
\def\gini{\texttt{gini}}
\def\equality{\texttt{eq}}
\def\productivity{\texttt{prod}}

\def\plannerpolicy{\pi_p}
\def\socialwelfare{\texttt{swf}}

\def\socialwelfareweight{\omega}

\def\figonea{a}
\def\figoneb{b}

\def\figtwoa{a}
\def\figtwob{b}
\def\figtwoc{c}

\def\figthreea{a}
\def\figthreeb{b}
\def\figthreec{c}
\def\figthreed{d}
\def\figthreee{e}
\def\figthreef{f}

\def\figfoura{a}
\def\figfourb{b}
\def\figfourc{c}
\def\figfourd{d}
\def\figfoure{e}

\def\figfourh{h}

\def\figfivea{a}
\def\figfiveb{b}
\def\figfivec{c}

%% file: src/v2/abstract.tex
\begin{sciabstract}
AI and reinforcement learning (RL) have improved many areas, but are not yet widely adopted in economic policy design, mechanism design, or economics at large.
At the same time, current economic methodology is limited by a lack of counterfactual data, simplistic behavioral models, and limited opportunities to experiment with policies and evaluate behavioral responses.
Here we show that machine-learning-based economic simulation is a powerful policy and mechanism design framework to overcome these limitations.
The AI Economist is a two-level, deep RL framework that trains both agents and a social planner who co-adapt, providing a tractable solution to the highly unstable and novel two-level RL challenge.
From a simple specification of an economy, we learn rational agent behaviors that adapt to learned planner policies and vice versa.
We demonstrate the efficacy of the AI Economist on the problem of optimal taxation.
In simple one-step economies, the AI Economist recovers the optimal tax policy of economic theory.
In complex, dynamic economies, the AI Economist substantially improves both utilitarian social welfare and the trade-off between equality and productivity over baselines. It does so despite emergent tax-gaming strategies, while accounting for agent interactions and behavioral change more accurately than economic theory.
These results demonstrate for the first time that two-level, deep RL can be used for understanding and as a complement to theory for economic design, unlocking a new computational learning-based approach to understanding economic policy.
\end{sciabstract}

%% file: src/v2/intro.tex
\hypertarget{introduction-700-words-currently-800}{
    \section{
        Introduction
    }\label{introduction-700-words-currently-800}
}
Economic policies need to be optimized to tackle critical global socio-economic issues and achieve social objectives.
For example, tax policy needs to balance equality and productivity, as large inequality gaps cause loss of economic opportunity~\autocite{inequality_matters_2013} and adverse health effects~\autocite{subramanian2004income}.
However, the problem of optimal policy design is very challenging, even when the policy objectives can be agreed upon.

Policy optimization poses a \emph{mechanism design}~\autocite{myerson1981optimal} problem: the government (\emph{social planner}) aims to find a policy under which the (boundedly) rational behaviors of affected economic agents yield the desired social outcome.\footnote{We only consider rational behaviors in this work, although our framework can be extended to include boundedly rational actors.}
Theoretical approaches to policy design are limited by analytical tractability and thus fail to capture the complexity of the real world.
Empirical studies are challenged by the lack of counterfactual data and face the \emph{Lucas critique}~\autocite{lucas1976econometric} that historical data do not capture behavioral responses to policy behavior.
Furthermore, opportunities for rigorous, real-world experimentation are limited and come with ethical questions~\autocite{rivlin1975ethical}.

Computational and machine learning techniques for \emph{automated} mechanism design
~\autocite{ConitzerS02, SandholmAutomatedDesign03, Narasimhan_ijcai16, baumann_adaptive_2018, dutting_optimal_2019}
show promise towards overcoming existing limitations, but a general computational framework for policy design remains lacking.
The challenge with policy design comes from needing to solve a highly non-stationary, \emph{two-level}, sequential decision-making problem where all actors (the agents and the government) are \emph{learning}: economic agents learn rational, utility-maximizing behaviors and the government learns to optimize its own objective via policy choices.

\paragraph{A New Machine Learning Challenge.}
Using deep reinforcement learning (RL) with multiple agents has been underexplored as a solution framework for mechanism design.
% Although mechanism design is well-motivated in the economic setting, it is an underexplored problem in reinforcement learning (RL) and, in particular, in the multi-agent setting.
%
Recent advances in deep RL have mostly studied the single-level setting; for example, state-of-the-art deep RL systems such as AlphaGo~\autocite{silver2017mastering} and AlphaStar~\autocite{vinyals2019grandmaster} optimized actors under fixed reward functions.
In contrast, in the two-level setting agents' effective reward functions depend on (changes in) the planner's policy, which leads to a highly unstable learning and co-adaptation problem.

Significant advances in multi-agent RL have focused on cooperative problems \autocite{vinyals2019grandmaster,OpenAI_dota}, and social dilemmas with fixed reward functions \autocite{leibo_multi-agent_2017}, but dynamical systems of heterogenous self-interested agents with changing incentives have been little studied at scale.

As such, few tractable computational learning approaches to mechanism design exist that scale to sequential settings with high-dimensional feature spaces.
Consequently, machine learning has so far not been widely applied to economic policy design.
In fact, more generally, economics as a field has not seen wide adoption of deep RL or related AI methods.

\paragraph{The AI Economist.}
Here we introduce the \emph{AI Economist}, a new and powerful framework that combines machine learning and AI-driven economic simulations to overcome the limitations faced by existing approaches.
Specifically, the AI Economist shows the efficacy and viability of using 1) AI-driven economic simulations and 2) two-level RL as a new paradigm for economic policy design.

\paragraph{AI-driven Simulations.}
We show that AI-driven simulations capture features of real-world economies without the need for hand-crafted behavioral rules or simplifications to ensure analytic tractability.
We use both a single-step economy and a multi-step, micro-founded economic simulation, \textit{\GatherTradeBuild{}}.
\GatherTradeBuild{} features multiple heterogenous economic agents in a two-dimensional spatial environment.
Productivity and income elasticity emerge as the result of the strategic behavior of multiple agents, rather than from statistical assumptions.
Moreover, \GatherTradeBuild{} includes trading between agents and simulates the economy over extended periods of time, i.e., spanning 10 tax periods, each of 100 days.
As such, the dynamics of \GatherTradeBuild{} are more complex than those considered in traditional tax frameworks and serve as a rich testbed for AI-driven policy design.

\paragraph{AI-driven Policy Design with Two-level, Deep RL.}
The AI Economist uses two-level, deep RL to learn optimal policies: at the level of individual agents within the economy and at the level of the social planner.
Both the agents and the social planner use deep neural networks to implement their policy model.
% The AI Economist uses two-level, deep RL to learn optimal policies in simulations with actors (worker agents and a social planner advising the government) who all learn.
%
Two-level RL compares the performance of billions of economic designs, making use of agents whose behaviors are learned along with the optimal planner policy.

Two-level RL is natural in many contexts, e.g., mechanism design, the principal-agent problem, or regulating systems with (adversarial) agents with misaligned or unethical incentives.
However, it poses a highly unstable learning problem, as agents need to continuously adapt to changing incentives.
The AI Economist solves the two-level problem through the use of \emph{learning curricula}~\autocite{bengio2009curriculum} and \emph{entropy-based regularization}~\autocite{williams1991function}, providing a tractable and scalable solution.
Our approach stablizes training using two key insights: (1) agents should not face significant utility costs that discourage exploration early during learning, and (2) the agents and social planner should be encouraged to gradually explore and co-adapt.

The AI Economist framework provides numerous advantages.

\begin{itemize}
\item It does not suffer from the Lucas critique. By design, it considers actors who co-adapt with economic policy.
\item Nor does it suffer the problems from using simulated agents with \emph{ad hoc} behavioral rules; rather, the use of RL provides rational agent behavior.
\item The simulation framework is flexible, supporting a configurable number of agents and various choices in regard to economic processes.
\item The designer is free to choose any policy objective and this does not have to be analytically tractable or differentiable.
\item The use of RL requires only observational data and does not require prior knowledge about the simulation or economic theory.
\end{itemize}

\paragraph{Optimal Tax Policy.}
We demonstrate the efficacy of the AI Economist on the problem of \textit{optimal tax policy design}~\autocite{diamond1971optimal,mirrlees_optimal_1976,mankiw_optimal_2009}, which aims to improve social welfare objectives, for example finding the right balance of equality and productivity.
In brief, tax revenue can be used to redistribute wealth, invest in infrastructure, or fund social programs.
At the same time, tax rates that are too high may disincentivize work and elicit strategic responses by tax-payers.

Theory-driven approaches to tax policy design have needed to make simplifications in the interest of analytical tractability~\autocite{RePEc:eee:pubchp:3-21}.
For example, typical models use static, one-step economies~\autocite{saez_using_2001,saez2016generalized} and make use of assumptions about people's sensitivity to tax changes (elasticity).
Although work in \emph{New Dynamic Public Finance (NDPF)}~\autocite{kocherlakota_new_2010,stantcheva2020} seeks to model multi-step economies, these models quickly become intractable to study analytically.
Concrete results are only available for two-step economies~\autocite{albanesi2006dynamic}.
These theoretical models also lack interactions between agents, such as market-based trading, and consider simple, inter-temporal dynamics.

Previous simulation work that makes use of agent-based modeling (ABM)~\autocite{bonabeau_agent-based_2002,bloomquist_tax_2011-1,miguel_exploring_2012,garrido_agent_2013,pennwhartonbudgetmodel,matchingirsstatistics2018} avoids problems of analytical tractability but uses complex and \emph{ad hoc} behavioral rules to study emergent behavior, this complicating the interpretation of results.
Moreover, the behavior of ABM agents is often rigid and lacking in strategic or adaptive behavior.

%%%%%%%%%%%%%%
%%%%%%%%%%%%%%
%%%%%%%%%%%%%%
\begin{figure}[t!]
\centering
\includegraphics[width=\linewidth]{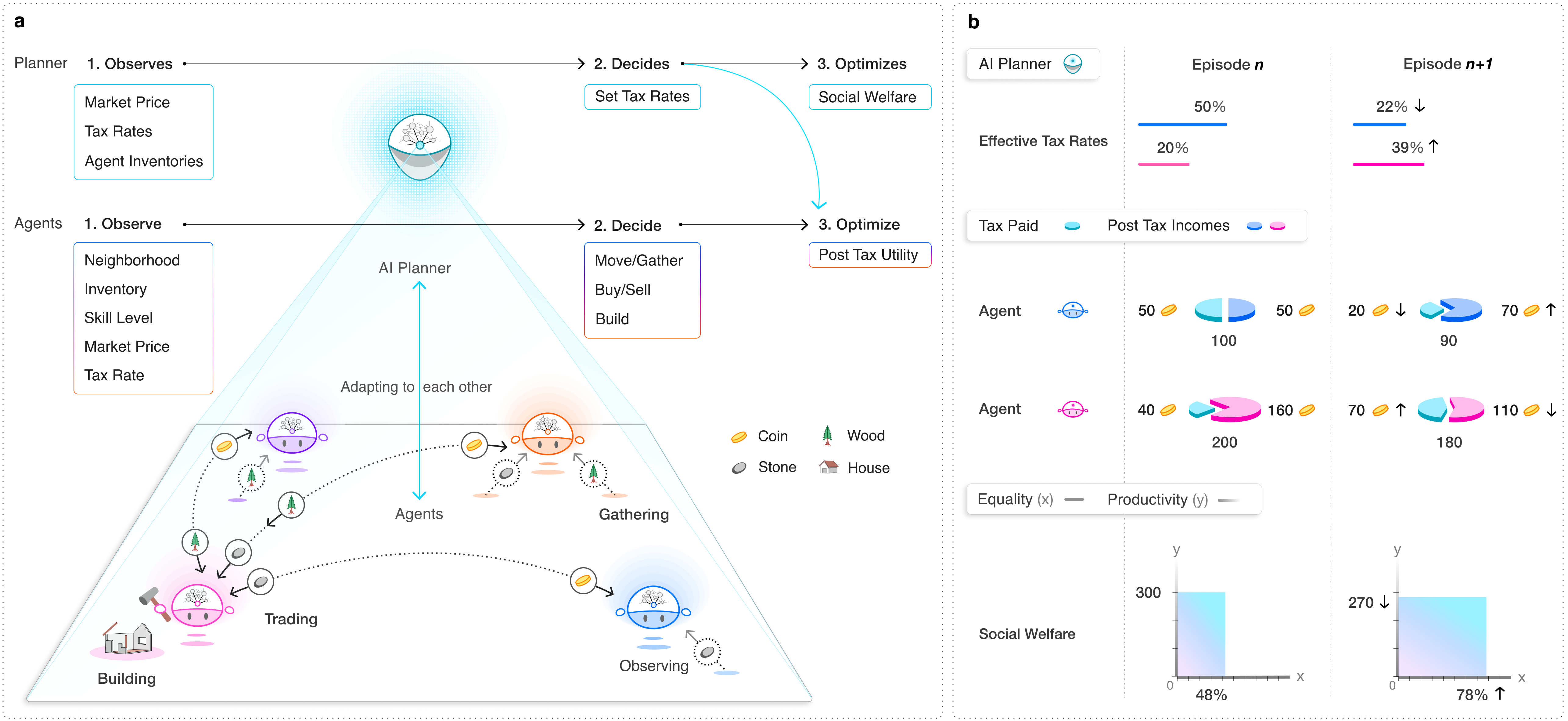}
\caption{
    \textbf{AI-driven economic simulations and two-level reinforcement learning (RL).}
    \textbf{\figonea,} An AI social planner optimizes social welfare by setting income tax rates in an economic simulation with AI agents.
    The agents optimize their individual post-tax utility by deciding how to perform labor and earn income.
    Both the planner and agents use RL to co-adapt and optimize their behavior.
    Agents need to optimize behavior in a non-stationary environment, as the planner's tax decisions change the reward that agents experience.
    \textbf{\figoneb,}
    Illustration of co-adaptation and two-level learning in an economy with two agents. Simulations proceed in episodes that last for 10 tax years, with 100 timesteps in each simulated year. During learning, between any episodes $n$ and $n+1$, the planner changes tax rates, which, after behavioral changes, leads to higher social welfare, here defined as the product of productivity and equality.
    }
    \label{fig:ai-driven-econ-sim-and-two-level-rl}
\end{figure}
%%%%%%%%%%%%%%
%%%%%%%%%%%%%%
%%%%%%%%%%%%%%

\paragraph{Experimental Validation.}
We provide extensive proof that the AI Economist provides a sound, effective, and viable approach to understanding, evaluating, and designing economic policy design.
%
% \art{TODO: rephrase, as with abstract.}
%
We study optimal tax design in a single-step economy and the multi-step \GatherTradeBuild{} environment, which implements a dynamic economy of heterogeneous, interacting agents that is more complex than the economic environments assumed in state-of-the-art tax models.
%
% \art{TODO: expand on GTB to drive home why it is interesting and meaningfully complex. Also use this text to define ``Open-Quadrant.''}
%
We show that the use of RL yields emergent agent behaviors that align well with economic intuition, such as specialization and tax gaming, phenomena that are not captured through analytical approaches to tax policy design.
This happens even with a small number of agents (4 and 10 agents in our experiments).

We show that policy models using two-level RL are effective, flexible, and robust to strategic agent behaviors through substantial quantitative and qualitative results:
\begin{itemize}
    \item In one-step economies, the AI Economist recovers the theoretically optimal tax policy derived by Saez \autocite{saez_using_2001}.
    This demonstrates the use of two-level RL is sound.

    \item In \GatherTradeBuild{} economies, tax policies discovered by the AI Economist provide a substantial improvement in social welfare for two different definitions of social welfare and in various spatial world layouts; e.g., in the Open-Quadrant world with four agents, \emph{utilitarian social welfare} increases by 8\%, and the \emph{trade-off between equality and productivity} increases by 12\% over the prominent Saez tax framework~\autocite{saez_using_2001}.

    \item In particular, AI social planners improve social welfare despite strategic behavior by AI agents seeking to lower their tax burden.

    \item AI-driven tax policies improve social welfare by using different kinds of tax schedules than baseline policies from economic theory.
    This demonstrates that analytical methods fail to account for all of the relevant aspects of an economy, while AI techniques do not require simplifying assumptions.

    \item Our work gives new economic insights: it shows that the well-established Saez tax model, while optimal in a static economy, is suboptimal in more realistic dynamic economies where it fails to account for interactions between agents.
    Our framework enables us to precisely quantify behavioral responses and agent interactions.
\end{itemize}

\paragraph{Ethical Disclaimer.}
As a point of caution, while the \GatherTradeBuild{} environments provide a rich testbed for demonstrating the potential of AI-driven simulation, they do not articulate the full range of economic opportunities, costs, and decisions faced by \emph{real-world individuals}, nor their distribution of relevant attributes.
More realistic AI-driven simulations are needed to support real-world policymaking, and defining the criteria for sufficient realism will require widespread consultation.
By extension, any conclusions drawn from experiments in these environments face the same limitations and, therefore, are not meant to be applied to any specific real-world economies.
See Section~\ref{Ethics-Discussion} for an extensive discussion on ethical risk.

%% file: src/v2/main_body.tex
\hypertarget{ai-driven-economic-simulations-300-words-265-currently}{
    \section{
        AI-driven Economic Simulations
    }\label{ai-driven-economic-simulations-300-words-265-currently}
}
%%%%%%%%%%%%%%%%
The AI Economist framework applies RL in two key ways: (1) to describe how \textit{rational} agents respond to alternative policy choices, and (2) to \textit{optimize} these policy choices in a principled economic simulation.
%%%%%%%%%%%%%%%%
Specifically, economic simulations need to capture the relevant economic drivers that define rational behavior.
%%%%%%%%%%%%%%%%
As such, a key strength of this framework is that finding rational behaviors along with an optimal policy remains tractable even with complex specifications of economic incentives and dynamics.

\paragraph{Simulation Dynamics.}
%%%%%%%%%%%%%%%%
We apply the AI Economist to the problem of optimal taxation (Figure \ref{fig:ai-driven-econ-sim-and-two-level-rl}).
%%%%%%%%%%%%%%%%
The set-up follows the Mirrleesian framework of non-linear optimal taxation subject to incentive constraints~\autocite{mirrlees_optimal_1976}. Here, the incentive constraints are represented through the rational behavior of agents, who optimize behavior subject to income tax and income redistribution.

Our simulations run for a finite number of timesteps $H$ and capture several key features of the Mirrleesian framework: that agents perform \textit{labor} $\labor$ in order to earn \textit{income} $\income$, where \textit{skill} determines how much income an agent earns for a given amount of labor; that an agent's utility increases with its \textit{post-tax} income and decreases with its labor; and that agents are heterogeneously skilled.

The simulation captures these concepts through its dynamics, i.e. the actions available to the actors and how those actions $\ac_t$ influence the world state $\st_t$ at timestep $t$.
%%%%%%%%%%%%%%%%
For example, agents may move spatially to collect resources, trade with one another, or spend resources to build houses; each such action accrues \emph{labor} but may generate \emph{income}, with higher \textit{skill} $\skill$ leading to higher incomes for the same actions.

\paragraph{Taxation.}
%%%%%%%%%%%%%%%%
Agents pay \emph{taxes} on the income they earn according to a tax schedule $\tax(\income, \mtaxrate)$, which determines taxes owed as a function of income and a set of \emph{bracketed marginal tax rates} $\mtaxrate$.
%%%%%%%%%%%%%%%%
The planner controls these tax rates, with all agents facing the same tax schedule, where this schedule can change at the start of each tax year.
%%%%%%%%%%%%%%%%
Collected taxes are evenly redistributed back to agents.
%%%%%%%%%%%%%%%%
For simplicity, we use fixed bracket intervals, and the planner only sets the marginal rates.

\paragraph{Behavioral Models.}
%%%%%%%%%%%%%%%%
Each actor (whether agent or planner) uses a deep neural network to encode its behavior as a probability distribution $\pi(\ac_t | \ob_t)$ over available actions, given observation $\ob_t$.
Following economic theory, each actor observes only a portion of the full world state $\st_t$.
For instance, the planner can observe trade activity but not an agent's skill level.
%%%%%%%%%%%%%%%%
Actors' objectives, i.e. post-tax utility for agents and social welfare for the planner, are captured in the \textit{reward function} used to train each behavioral policy $\policy$.
In this way, the AI Economist uses RL to describe rational agent behavior and optimize policy choices in complex, sequential economies beyond the reach of traditional analysis.

%%%%%%%%%%%%%%%%
%%%%%%%%%%%%%%%%
%%%%%%%%%%%%%%%%
\begin{algorithm}[ht]
    \floatname{algorithm}{Algorithm}
    \caption{
        \textbf{Two-level Reinforcement Learning.} Agents and social planner learn simultaneously. Bold-faced symbols indicate quantities for multiple agents. Note that agents share weights.}
    \label{ext-data-alg:two-level-rl}
    \begin{algorithmic}
        \Input
        \Desc{$\samplinghorizon$}{Sampling horizon}
        \Desc{$\taxperiodlen$}{Tax period length}
        \Desc{$\sA$}{On-policy learning algorithm (in this work, PPO~\autocite{schulman2017proximal})}
        \Desc{$\stoppingcriterion$}{Stopping criterion (for instance, agent and planner rewards have not improved)}
        \EndInput
        \Output
        \Desc{$\theta$}{Trained agent policy weights}
        \Desc{$\phi$}{Trained planner policy weights}
        \EndOutput
        \State $\st, \V{\ob}, \ob_p, \V{\hidden}, \hidden_p \Leftarrow \st_0, \V{\ob}_0, \ob_{p, 0}, \V{\hidden}_0, \hidden_{p, 0}$ \Comment{Reset episode: initialize world state $\st$, observation $\ob$, hidden states $\hidden$}
        \State $\theta, \phi \Leftarrow \theta_0, \phi_0$ \Comment{Initial agent and planner policy weights}
        \State $\dataset, \dataset_p \Leftarrow \{\}, \{\}$ \Comment{Reset agent and planner transition buffers}
        \While{training}
        \For{$t = 1, \ldots, \samplinghorizon$}
            \State $\Ac, \V{\hidden} \Leftarrow \V{\policy}(\cdot | \V{\ob}, \V{\hidden}, \mweight)$ \Comment{Sample agent actions; update hidden state}
            \If{$t$ mod $\taxperiodlen$ = 0}  \Comment{First timestep of tax period}
                \State $\mtaxrate, \hidden_p \Leftarrow \plannerpolicy(\cdot | \ob_p, \hidden_p, \pweight)$ \Comment{Sample marginal tax rates; update planner hidden state}
            \Else
                \State $\texttt{no-op}, \hidden_p \Leftarrow \plannerpolicy(\cdot | \ob_p, \hidden_p, \pweight)$ \Comment{Only update planner hidden state}
            \EndIf
            \State $\st', \V{\ob}', \ob_p', \V{\rew}, \rew_p \Leftarrow \texttt{Env.step}(\st, \Ac, \mtaxrate)$  \Comment{Next state / observations, pre-tax reward, planner reward}
            \If{$t$ mod $\taxperiodlen$ = $\taxperiodlen$ - 1}  \Comment{Last timestep of tax period}
                \State $\st', \V{\ob}', \ob_p', \V{\rew}, \rew_p \Leftarrow \texttt{Env.tax}(\st', \mtaxrate)$  \Comment{Apply taxes; compute post-tax rewards}
            \EndIf
            \State $\dataset \Leftarrow \dataset \cup \{(\V{\ob}, \Ac, \V{\rew}, \V{\ob}')\}$ \Comment{Update agent transition buffer}
            \State $\dataset_p \Leftarrow \dataset_p \cup \{(\ob_p, \mtaxrate, \rew_p, \ob'_p)\}$ \Comment{Update planner transition buffer}
            \State $\st, \V{\ob}, \ob_p \Leftarrow \st', \V{\ob}', \ob'_p$
        \EndFor
        \State Update $\mweight, \pweight$ using data in $\dataset, \dataset_p$ and $\sA$.
        \State $\dataset, \dataset_p \Leftarrow \{\}, \{\}$ \Comment{Reset agent and planner transition buffers}
        \If{episode is completed}
            \State $\st, \V{\ob}, \ob_p, \V{\hidden}, \hidden_p \Leftarrow \st_0, \V{\ob}_0, \ob_{p, 0}, \V{\hidden}_0, \hidden_{p, 0}$ \Comment{Reset episode}
        \EndIf
        \If{criterion $\stoppingcriterion$ is met}
            \Return $\theta, \phi$
        \EndIf
        \EndWhile
    \end{algorithmic}
\end{algorithm}
%%%%%%%%%%%%%%%%
%%%%%%%%%%%%%%%%
%%%%%%%%%%%%%%%%

\hypertarget{two-level-reinforcement-learning-400-words-currently-451}{
    \section{
        Two-Level Reinforcement Learning
    }\label{two-level-reinforcement-learning-400-words-currently-451}
}
%%%%%%%%%%%%%%%%
%%%%%%%%%%%%%%%%
Under the AI Economist framework, all actors (i.e. the AI agents and the AI planner) learn and adapt using RL~\autocite{sutton2018reinforcement}, see \ExtendedDataString{}Algorithm \ref{ext-data-alg:two-level-rl}.
%%%%%%%%%%%%%%%%
Each actor learns a behavioral policy $\pi$ to maximize its objective (expected sum of future rewards).
%%%%%%%%%%%%%%%%
Each actor also learns a \emph{value function}, which estimates this expectation given observation $\ob_t$.
%%%%%%%%%%%%%%%%
Actors iteratively \emph{explore} actions by sampling from their current behavioral model, and improve this model across episodes by training on experiential data.
%%%%%%%%%%%%%%%%
RL agents can be optimized for any reward function and this does not have to be analytical.

An agent $i$ maximizes \emph{expected total discounted utility}:
%%%%%%%%%%%%%%%%
\eq{\label{eq:agent-utility}
    %%%%%%%%%%%%%%%%
    \max_{\policy_i}
    %%%%%%%%%%%%%%%%
    \E_{\ac_i \sim \policy_i, \Ac_{-i} \sim \bm{\policy}_{-i}, \st'\sim \trans}
    \brcksq{
        \left.\sum_{t=1}^H \gamma^t \rew_{i,t} + \util_{i,0}\right| \mtaxrate
    }, \quad
    %%%%%%%%%%%%%%%%
    \rew_{i,t} = \util_{i,t} - \util_{i,t-1},
}
%%%%%%%%%%%%%%%%
given tax rates $\mtaxrate$, discount factor $\df$, and utility $\util_{i, t}$. Here $\st'$ is the state following $\st$, and $\trans$ represents the simulation dynamics.
%%%%%%%%%%%%%%%%
We use {\em isoelastic utility}~\autocite{arrow1971theory}:
%%%%%%%%%%%%%%%%
\eq{\label{eq:crra-utility}
    \util_{i,t} = \frac{\money_{i,t}^{1 - \eta} - 1}{1 - \eta} - \totallabor_{i, t},\quad \eta > 0,
}
%%%%%%%%%%%%%%%%
which models diminishing marginal utility over money endowment $\money_{i,t}$, controlled by $\eta > 0$, and the linear disutility of total labor $\totallabor_{i, t}$.
%%%%%%%%%%%%%%%%
The planner maximizes expected social welfare:
%%%%%%%%%%%%%%%%
\eq{\label{eq:social-planner-objective}
\max_{\plannerpolicy}
\E_{
    \mtaxrate \sim \plannerpolicy,
    \Ac \sim \bm{\policy},
    \st' \sim \trans
    }
    \brcksq{
        \sum_{t=1}^H \df^t \rew_{p,t} + \socialwelfare_0
    }, \quad
    %%%%%%%%%%%%%%%%
    \rew_{p,t} = \socialwelfare_t - \socialwelfare_{t-1},
}
where $\socialwelfare_t$ is social welfare at time $t$.
%%%%%%%%%%%%%%%%
We take $\socialwelfare$ as a utilitarian objective (an average of all agent utilities weighted by their inverse pre-tax income), or alternatively as the product of equality and productivity (representing a balance between equality and productivity).
%%%%%%%%%%%%%%%%
For details, see Methods.

Agents need to adapt to policies set by the planner, and vice versa (Figure \ref{fig:ai-driven-econ-sim-and-two-level-rl}).
%%%%%%%%%%%%%%%%
This is a challenging non-stationary learning problem.
%%%%%%%%%%%%%%%%
While learning, the planner in effect adjusts agent reward functions because taxes influence the post-tax income that agents receive as a result of payments and redistributions.
%%%%%%%%%%%%%%%%
As the tax schedule changes, the optimal behavior for agents changes.
%%%%%%%%%%%%%%%%
This instability is exacerbated by mutual exploration.

These challenging learning dynamics reflect the nested optimization problem that two-level RL attempts to solve.
%%%%%%%%%%%%%%%%
That is, we aim to find the tax rates that maximize social welfare, subject to the constraint that agents' behaviors maximize their own utility given the tax rates.
%%%%%%%%%%%%%%%%
Planner learning (the outer level) serves to maximizing social welfare, whereas agent learning (the inner level) serves to ensure that the constraint is satisfied.
%%%%%%%%%%%%%%%%
Our approach to two-level RL follows from the intuition that instability depends on how well the agent-optimality constraint is satisfied during learning.

To stabilize learning, our approach combines two key ideas: \emph{curriculum learning}~\autocite{bengio2009curriculum} and \emph{entropy regularization}~\autocite{williams1991function}.
%%%%%%%%%%%%%%%%
This effectively \textit{staggers} agent and planner learning such that agents are well-adapted to a wide range of tax settings before the learning of the planner begins.
%%%%%%%%%%%%%%%%
In particular, we use the early portion of training to gradually introduce labor costs and, later, taxes.
%%%%%%%%%%%%%%%%
These curricula are based on the key intuition that suboptimal agent strategies may incur punitively high cost of labor and taxes, while earning insufficient income to yield positive utility, and this may discourage RL agents from continuing to learn.
%%%%%%%%%%%%%%%%
%%%%%%%%%%%%%%%%
%%%%%%%%%%%%%%%%
We schedule the entropy regularization applied to $\plannerpolicy$ such that agents are initially exposed to highly random taxes.
%%%%%%%%%%%%%%%%
Random taxes provide the training experience needed for agent policies to appropriately condition actions on the observed tax rates, for a wide range of possible taxes.
%%%%%%%%%%%%%%%%
As described above, this is an important precondition for stably introducing planner optimization.
%%%%%%%%%%%%%%%%
Lastly, the entropy of policy models are strongly regularized to encourage exploration and gradual co-adaptation between the agents and social planner throughout the remainder of training.
%%%%%%%%%%%%%%%%
For details, see Methods.

We note that unlike previous strategies for overcoming instability in multi-agent RL~\autocite{foerster_learning_2017,lowe_multi-agent_2017}, ours is tailored to the nested optimization intrinsic to the two-level setting.

\hypertarget{matching-analytical-results-for-one-step-economies-200-words-currently-230}{
    \section{
        Validation in a One-Step Economy
    }\label{matching-analytical-results-for-one-step-economies-200-words-currently-230}
}
%%%%%%%%%%%%%%%%
The most prominent solution for optimal taxation is the analytical framework developed by Saez~\autocite{saez_using_2001}.
%%%%%%%%%%%%%%%%
This framework analyzes a simplified model where both the planner and the agents each make a single decision: the planner setting taxes and the agents choosing labor.
%%%%%%%%%%%%%%%%
This analysis describes the welfare impact of a tax rate change via its mechanical effect on redistribution and its behavioral effect on the underlying income distribution.
%%%%%%%%%%%%%%%%
%%%%%%%%%%%%%%%%
The resulting formula computes theoretically optimal tax rates as a function of the income distribution and the \emph{elasticity} of income with respect to the marginal tax rate.\footnote{
In practice, these income elasticities typically need to be estimated from empirical data, which is a non-trivial task~\autocite{gruber2002elasticity}.}

We first validate our approach in these simplified one-step economies.
%%%%%%%%%%%%%%%%
Each agent chooses an amount of labor that optimizes its post-tax utility, and this optimal labor depends on its skill and the tax rates, and it does not depend on the labor choices of other agents.
%%%%%%%%%%%%%%%%
Before the agents act, the planner sets the marginal tax rates in order to optimize social welfare, taken here to be utilitarian.

\begin{figure}[t!]
    \centering
    \includegraphics[width=\linewidth]{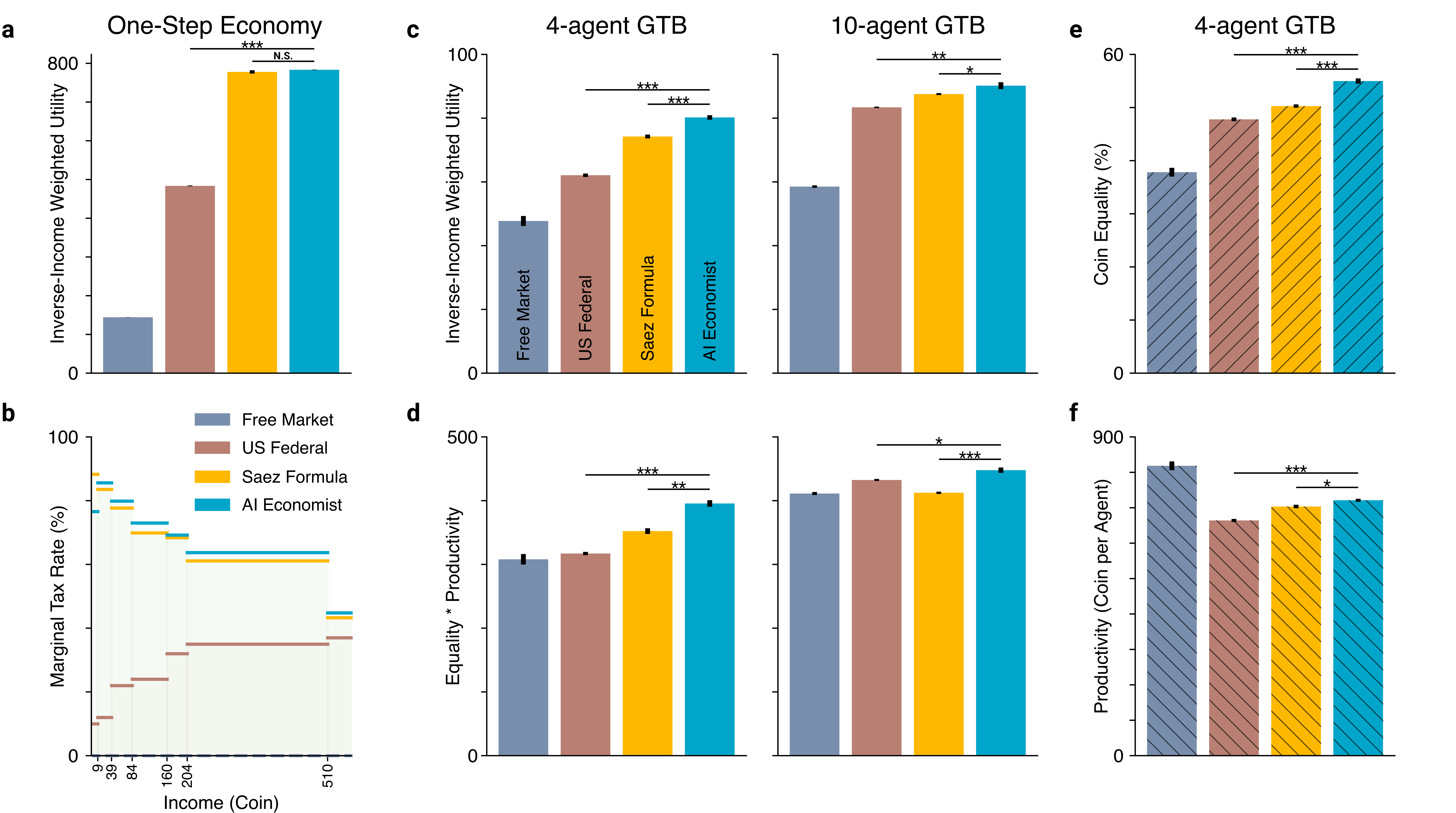}
    \caption{
        \textbf{Quantitative results in a one-step economy and the Open-Quadrant \GatherTradeBuild{} environment.}
        \textbf{\figthreea-\figthreeb,} The results of the AI Economist and the  Saez tax  are highly consistent in the one-step economy, both in terms of utilitarian social welfare (\figthreea) and the tax schedule (\figthreeb).
        %%%%%%%%%%%%%%%%
        \textbf{\figthreec-\figthreed,} In the \GatherTradeBuild{} environment (GTB) with 4 and 10 agents, the AI Economist outperforms baselines when optimizing the utilitarian social welfare objective (\figthreec) and when optimizing the equality-times-productivity objective (\figthreed).
        \textbf{\figthreee-\figthreef,} Overall coin equality (\figthreee) and average productivity (\figthreef) achieved by each tax model in the 4-agent Open Quadrant scenario.
        Each bar represents the average end-of-training metrics over 10 random seeds (5 for the one-step economy), with error bars denoting standard error. %*$p<0.05$, **$p<0.001$, ***$p<0.00001$.
        Asterisks indicate a statistically significant difference at an $\alpha$ level of 0.05 (*), 0.001 (**), or 0.00001 (***).
        N.S. denotes not statistically significant ($p > 0.05$).
        All social welfare, productivity, and equality differences between the AI Economist and baselines are statistically significant, except for the difference in social welfare between the AI Economist and the Saez tax in the one-step economy (\figfivea).
    }
    \label{fig:quantitative-results}
\end{figure}

We compare the economy under the \emph{Saez tax}, and  the AI Economist.
%%%%%%%%%%%%%%%%
In both cases, AI agents learn to optimize their own utility given their tax setting.
%%%%%%%%%%%%%%%%
The Saez tax baseline computes tax rates based on our implementation of the Saez formula (induced through an optimal elasticity parameter found via grid-search, as detailed in the Methods), and the AI Economist learns tax rates via two-level RL.
%%%%%%%%%%%%%%%%
We include two additional baseline tax models here and throughout this work: the \emph{free market} (no taxes) and a stylized version of the \emph{US Federal} progressive tax schedule (see Methods for details).
%%%%%%%%%%%%%%%%
There is no \textit{a priori} expectation that either of the additional baselines should maximize social welfare; rather, they provide useful comparison and help to characterize behavioral responses to different tax choices.
%%%%%%%%%%%%%%%%
The AI Economist and the Saez tax schedule produce highly consistent tax schedules and social welfare, as shown in Figure \ref{fig:quantitative-results}\figthreea-\figthreeb.
%%%%%%%%%%%%%%%%
In comparison, the free market and US Federal achieve substantially worse social welfare.
%%%%%%%%%%%%%%%%
These results show that the AI Economist can reproduce optimal tax rates in economies that satisfy the simplifying assumptions of optimal tax theory and validate the soundness of our learning-based approach.

%%%%%%%%%%%%%%%%
%%%%%%%%%%%%%%%%
\begin{figure}[t!]
    \centering
    \includegraphics[width=\linewidth]{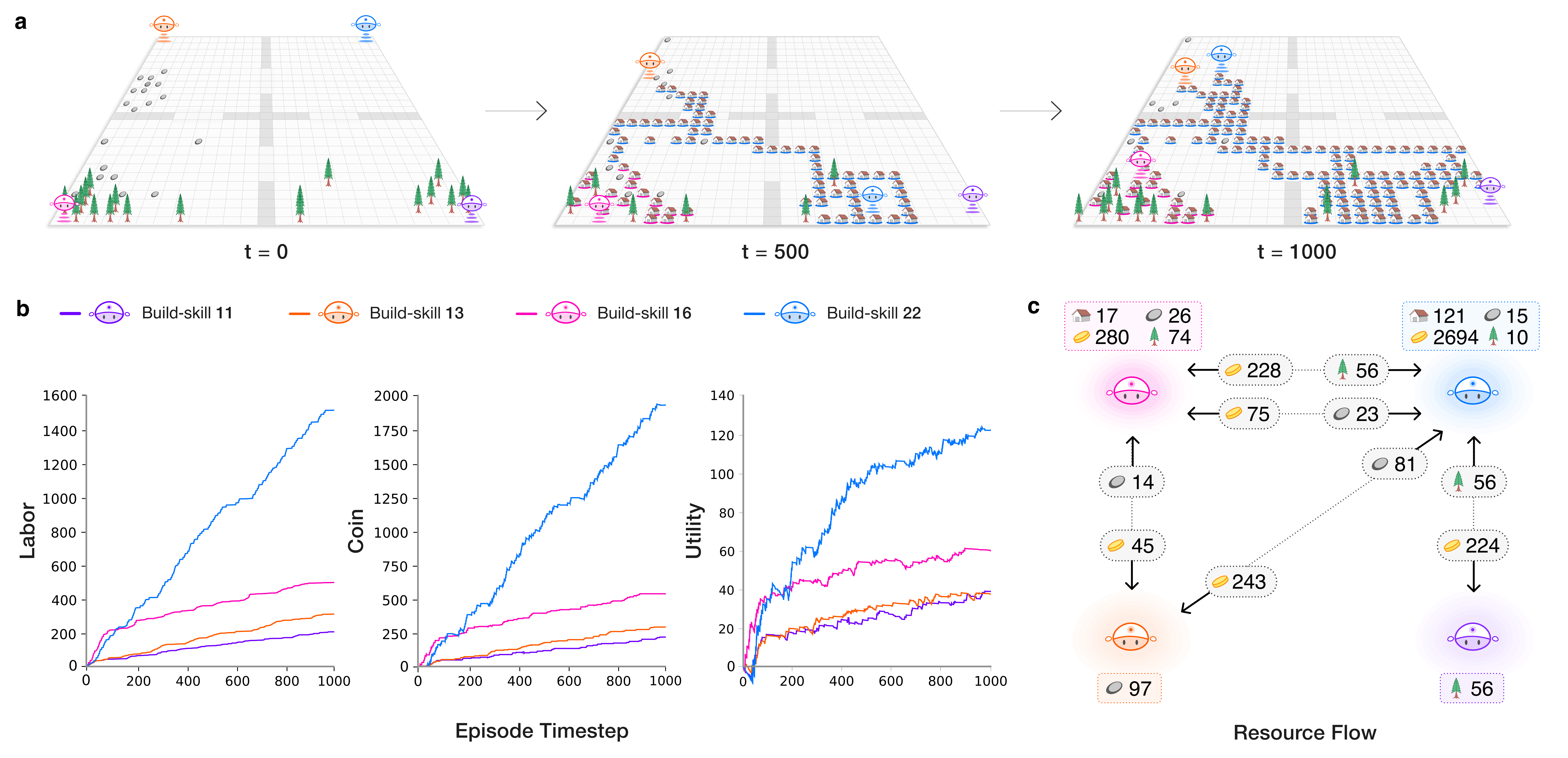}
    \caption{
        \textbf{Emergent phenomena in AI-driven economic simulations under the free market.}
        \textbf{\figtwoa,} Visualization of the spatial state of the world at $t = 0$, $500$, and $1000$ of an example episode in the 4-agent Open-Quadrant \GatherTradeBuild{} scenario. Agents specialize as \textit{builders} (blue agent) or \textit{gatherers} (others) depending on their build-skill.
        \textbf{\figtwob,} Labor, income, and utility over the course of the episode for all agents. Each quantity increases with build-skill in this setting. The highest build-skill (blue) agent chooses to do the most work, and earns larger income and ultimately experience the most utility.
        \textbf{\figtwoc,} Net resource flow between agents during the episode. The box adjacent to each agent show the resources it gathered and the coin it earned from building. Arrows between agents denote coin and resources exchanged through trading.
    }
    \label{fig:emergent-phenomena}
\end{figure}
%%%%%%%%%%%%%%%%
%%%%%%%%%%%%%%%%

\hypertarget{gather-trade-build-a-dynamic-economy}{
    \section{
        Gather-Trade-Build: a Dynamic Economy
    }\label{gather-trade-build-a-dynamic-economy}
}
%%%%%%%%%%%%%%%%
We study the \emph{\GatherTradeBuild{}} economy, a two-dimensional, spatiotemporal economy with agents who move, gather resources (stone and wood), trade, and build houses.
%%%%%%%%%%%%%%%%
\GatherTradeBuild{} captures the fundamental trade-off between equality and productivity intrinsic to optimal tax design (see below), and is a rich testbed to demonstrate the advantages of AI-driven policy design.

Each simulation simulates 10 tax years. Each tax year lasts 100 timesteps (so that $H=1000$), with the agents acting each timestep, and the planner setting and changing tax rates at the start of each tax year. The \GatherTradeBuild{} environment is depicted in Figure \ref{fig:ai-driven-econ-sim-and-two-level-rl}. For details, see Methods.

\paragraph{AI-driven Simulations Capture Macro-Economic Phenomena.}
%%%%%%%%%%%%%%%%
A key advantage is that AI-driven simulations capture macro-level features of real economies that are emergent purely through learned rational behavior and without being manually implemented.
%%%%%%%%%%%%%%%%
To illustrate this, we showcase three examples of AI-driven emergent behavior.

\paragraph{Example 1: Emergent Specialization.}
Each agent varies in its skill level.
%%%%%%%%%%%%%%%%
We instantiate this in our simulation as \textit{build-skill}, which sets how much income an agent receives from building a house.
%%%%%%%%%%%%%%%%
Build-skill is distributed according to a Pareto distribution.
%%%%%%%%%%%%%%%%
As a result, we observe that utility-maximizing agents \emph{learn to specialize} their behavior based on their build-skill, see Figure \ref{fig:emergent-phenomena}.
%%%%%%%%%%%%%%%%
Agents with low build-skill become \emph{gatherers}: they earn income by gathering and selling resources.
%%%%%%%%%%%%%%%%
Agents with high build-skill become \emph{builders}: they learn that it is more profitable to buy resources and then build houses.
%%%%%%%%%%%%%%%%
This emergent behavior is entirely due to \emph{heterogeneous} their experienced utility for different economic activity, and not due to fixed behavioral rules as in most traditional agent-based modeling.

\begin{figure}[t!]
    \centering
    \includegraphics[width=\linewidth]{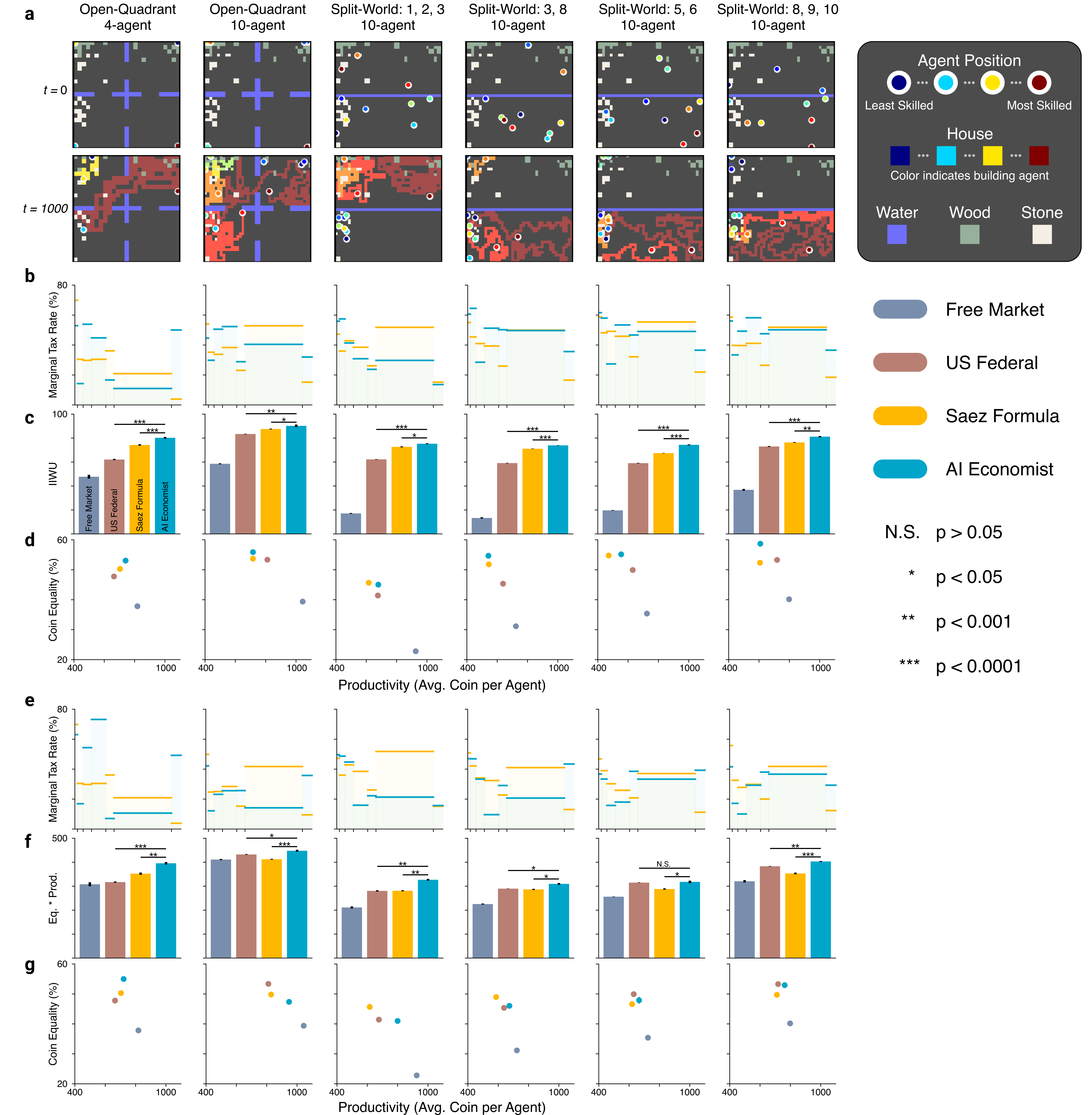}
    \caption{Description on the next page.}
    \label{ext-data-fig:gtb-quantative-results}
\end{figure}

\paragraph{Example 2: Equality-Productivity Trade-off.}
%%%%%%%%%%%%%%%%
Our AI simulations capture the trade-off between equality and productivity: as tax rates increase, equality increases through wealth transfers, but productivity falls as agents are less incentivized to work due to lower post-tax incomes (Figure \ref{fig:quantitative-results} and \ExtendedDataString{}Figure \ref{ext-data-fig:gtb-quantative-results}).
%%%%%%%%%%%%%%%%
As a demonstration of this, the free market (no tax) baseline always yields the highest productivity and lowest equality compared to the alternative tax models.
%%%%%%%%%%%%%%%%
Unlike standard theoretical models that rely on elasticity assumptions to capture this trade-off, we observe it as an emergent consequence of rational behavior.
%%%%%%%%%%%%%%%%

\paragraph{Example 3: AI Tax Gaming Strategies.}
%%%%%%%%%%%%%%%%
Our AI simulations yield emergent strategic behaviors.
%%%%%%%%%%%%%%%%
High-income agents learn to avoid taxes by moving labor and thus income between tax years in order to move more income to low-rate brackets.
%%%%%%%%%%%%%%%%
This can reduce the overall tax paid in comparison to earning a constant amount each year (Figure \ref{fig:economic-interactions-and-income-elasticity}c).
%%%%%%%%%%%%%%%%
Given the complexity of \GatherTradeBuild{} and similar dynamic economic environments, it is prohibitively complex for theory-driven methods to derive such temporal behavioral strategies.

Together, these examples show that AI-driven simulations capture features of real-world economies, purely through RL.
%%%%%%%%%%%%%%%%
Hence, AI-driven simulations provide a rich class of environments for policy design, unconstrained by analytic tractability.

%%%%%%%%%%%%%%%%
%%%%%%%%%%%%%%%%
%%%%%%%%%%%%%%%%
%%%%%%%%%%%%%%%%
%%%%%%%%%%%%%%%%
\addtocounter{figure}{-1}
\begin{figure}[t!]
    \caption{\textbf{
        Comprehensive quantitative results in the \GatherTradeBuild{} environment with the utilitarian or equality-times-productivity planner objective, across all settings: Open-Quadrant and 4 Split-World scenarios; 4 and 10 agents.
        }
        The AI Economist achieves significantly higher social welfare than all baselines.
        \textbf{a,} Spatial layouts of the Open-Quadrant and Split-World scenarios at the start ($t=0$) and end ($t=1000$) of example episodes.
        \textbf{b,} Tax schedules for the Saez tax (yellow) and the AI Economist (blue).
        \textbf{c,} Utilitarian social welfare objective (inverse-income weighted utility, labeled ``IIWU'') for all planners.
        \textbf{d,} Equality and productivity for all planners.
        For the data in b-d, the AI Economist is trained to maximize the utilitarian social welfare objective, and the Saez taxes use the best-performing elasticity for the utilitarian objective.
        \textbf{e-g,} As b-d, but for the data in e-g the AI Economist is trained to maximize the equality-times-productivity social welfare objective, and the Saez taxes use the best-performing elasticity for this objective, which is shown in f.
        Bars and dots represent the average end-of-training metrics over 10 (5) random seeds for the Open-Quadrant (Split-World) scenarios, with error bars denoting standard error.
        Asterisks indicate a statistically significant difference at an $\alpha$ level of 0.05 (*), 0.001 (**), or 0.00001 (***).
        N.S. denotes not statistically significant ($p > 0.05$).
        All social welfare differences between the AI Economist and baselines are statistically significant, except for the difference in equality-times-productivity (f) between the AI Economist and the US Federal tax in the \textit{Split-World-5,6} scenario.
    }
\end{figure}
%%%%%%%%%%%%%%%%
%%%%%%%%%%%%%%%%
%%%%%%%%%%%%%%%%
\hypertarget{ai-driven-optimal-taxation}{
    \section{
        AI-Driven Optimal Taxation
    }\label{ai-driven-optimal-taxation}
}
%%%%%%%%%%%%%%%%
We evaluate the AI Economist across different \GatherTradeBuild{} economies to validate that AI-driven policy design is effective, can be applied to different economic environments, and adapts to strategic behavior more successfully than baseline tax policies.

\paragraph{Settings.}
%%%%%%%%%%%%%%%%
We use two spatial layouts, \emph{Open-Quadrant} and \emph{Split-World}, each with different physical barrier placements and different agent starting positions.
%%%%%%%%%%%%%%%%
\emph{Open-Quadrant} features four areas laid out in a $2\times 2$ pattern, each area having a connection with its neighbor to allow agents to move between areas.
%%%%%%%%%%%%%%%%
\emph{Split-World} features two halves, separated by an impassable water barrier. This prevents agents from moving between the top and bottom halves of the map, which blocks agents from directly accessing certain resources.
%%%%%%%%%%%%%%%%

We consider four Split-World scenarios, each with 10 agents but differing in the subset of agents assigned to the resource-rich half.
%%%%%%%%%%%%%%%%
We consider two Open-Quadrant scenarios, with 4 agents in one version and 10 agents in the other.
%%%%%%%%%%%%%%%%
All 6 scenarios are illustrated in \ExtendedDataString{}Figure~\ref{ext-data-fig:gtb-quantative-results}a.
%%%%%%%%%%%%%%%%
For ease of exposition, we focus our fine-grained analyses on results in the 4-agent Open-Quadrant scenario.

%%%%%%%%%%%%%%%%
%%%%%%%%%%%%%%%%
%%%%%%%%%%%%%%%%
%%%%%%%%%%%%%%%%
%%%%%%%%%%%%%%%%

\paragraph{Improved Social Welfare.}
%%%%%%%%%%%%%%%%
As with the one-step economy, we compare the AI Economist against the \textit{free market}, \textit{US Federal}, and \textit{Saez tax} baselines across all of these settings (see Methods).
%%%%%%%%%%%%%%%%
The AI Economist achieves the highest social welfare throughout.
%%%%%%%%%%%%%%%%
%%%%%%%%%%%%%%%%
The combined results of these experiments are presented in \ExtendedDataString{}Figure \ref{ext-data-fig:gtb-quantative-results}.
%%%%%%%%%%%%%%%%
In the \emph{Open-Quadrant} layout with four (ten) agents (Figure \ref{fig:quantitative-results}), AI-driven taxes improve the utilitarian objective by over
%%%%%%%%%%%%%%%%
8\% (2\%)
%%%%%%%%%%%%%%%%
and the product of equality and productivity by over
%%%%%%%%%%%%%%%%
12\% (8.6\%)
%%%%%%%%%%%%%%%%
over the Saez tax.
%%%%%%%%%%%%%%%%
%%%%%%%%%%%%%%%%

We observe that the relative performance of the baselines depends on the choice of social welfare objective: the utilitarian objective is always higher when using the Saez tax compared to the US Federal tax; however, the opposite is often true for the equality-times-productivity objective (especially in settings with 10 agents).
%%%%%%%%%%%%%%%%
In contrast, the AI Economist is not tailored towards a particular definition of social welfare and flexibly adjusts its tax schedule to optimize the chosen objective, yielding the best social welfare throughout.

These results show the AI Economist is flexible, maintains performance with more agents, can be successfully optimized for distinct objectives, and works well in the face of adaptive, strategic behavior.

\begin{figure}[t!]
    \centering
    \includegraphics[width=\linewidth]{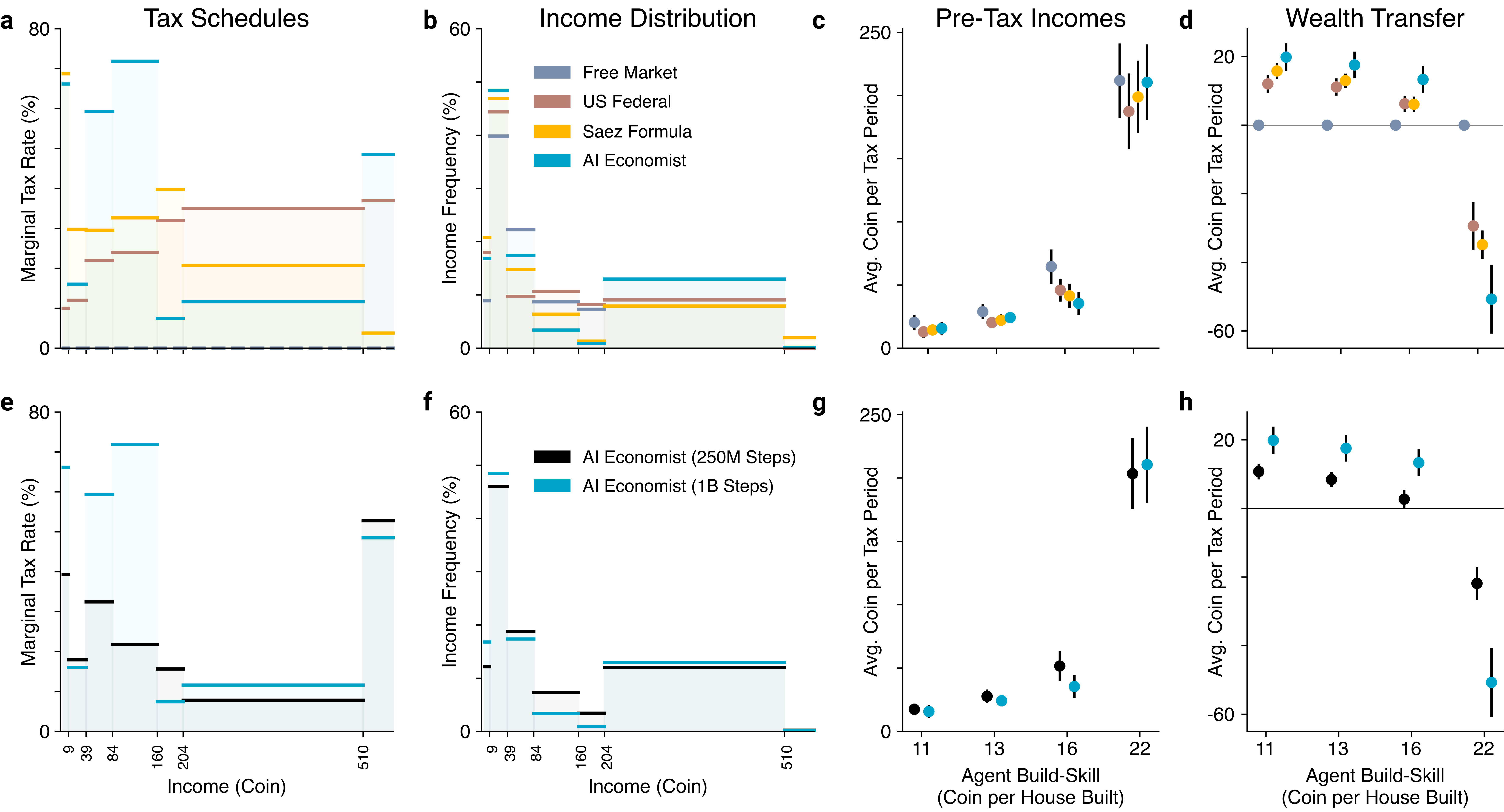}
    \caption{
        \textbf{Comparison of tax policies in the 4-agent Open-Quadrant \GatherTradeBuild{} environment.}
        \textbf{\figfoura,} Average marginal tax rates within each tax bracket.
        \textbf{\figfourb,} Frequency with which agent incomes fall within each bracket.
        \textbf{\figfourc,} Average pre-tax income of each agent (sorted by build-skill) under each of the tax models.
        \textbf{\figfourd,} Average wealth transfer resulting from taxation and redistribution.
        \textbf{\figfoure-\figfourh,} Same as \figfoura-\figfourd, comparing the AI Economist from early during training (250 million training samples) versus at the end of training (1 billion training samples).
        %%%%%%%%%%%%%%%%
        Dots denote averages and error bars denote standard deviation across episodes.
    }
    \label{fig:ai-tax-policy-wealth-transfer-and-strategies}
\end{figure}

\paragraph{Adaptation During Training.}
%%%%%%%%%%%%%%%%
During training, the AI Economist increases rates on the first (incomes of $0$ to $9$), third ($39$ to $84$), and fourth ($84$ to $160$) brackets, maintaining low rates otherwise, see Figure \ref{fig:ai-tax-policy-wealth-transfer-and-strategies}.
%%%%%%%%%%%%%%%%
%%%%%%%%%%%%%%%%
%%%%%%%%%%%%%%%%
This does not significantly shift the pre-tax income distribution, while the post-tax income distribution becomes more equal.
%%%%%%%%%%%%%%%%
The resulting tax schedule is distinctly different from the baselines, which use either increasing (progressive) or decreasing (regressive) schedules (Figure \ref{fig:ai-tax-policy-wealth-transfer-and-strategies}a).
%%%%%%%%%%%%%%%%
The AI Economist is neither: on average, it sets the highest marginal rates for incomes between $39$ and $160$ coins and the lowest rates for the adjacent brackets ($9$ to $39$ and $160$ to $510$ coins).
%%%%%%%%%%%%%%%%
Under the AI Economist, the low build-skill agents earn 9\% more from trading (Figure \ref{fig:economic-interactions-and-income-elasticity}\figfiveb),
%%%%%%%%%%%%%%%%
wealth transfers from the highest build-skill agent to others are 46\% larger (Figure \ref{fig:ai-tax-policy-wealth-transfer-and-strategies}d),
%%%%%%%%%%%%%%%%
income equality is at least 9\% higher (Figure \ref{fig:quantitative-results}e),
%%%%%%%%%%%%%%%%
the number of incomes in the second-to-highest bracket ($204$ to $510$ coins) is at least 64\% higher,
%%%%%%%%%%%%%%%%
and, 92\% smaller for the top bracket,
%%%%%%%%%%%%%%%%
%%%%%%%%%%%%%%%%
compared to baselines (Figure \ref{fig:ai-tax-policy-wealth-transfer-and-strategies}b).
%%%%%%%%%%%%%%%%
These numbers are measured over the last 400 episodes within each experiment group, which amounts to 4000 total tax periods and 16000 total incomes per group.

\paragraph{Behavior of Learned AI Tax Policies.}
%%%%%%%%%%%%%%%%
The AI Economist adapts to different environments: \ExtendedDataString{}Figure \ref{ext-data-fig:gtb-quantative-results} shows that the best-performing AI taxes behave differently across scenarios.

For instance, in the Open-Quadrant, the AI tax schedules are similar when optimizing for the two different social welfare objectives with 4 agents but this pattern changes with 10 agents, where objective-specific tax schedules emerge.
%%%%%%%%%%%%%%%%
Tax rates for the brackets between $9$ and $160$ coins follow different patterns, for example, and overall tax rates are lower when optimizing for equality times productivity.

Furthermore, in the Split-World, the AI tax schedule depends on which agents are in the resource-rich top half of the environment.
%%%%%%%%%%%%%%%%
As an example, when optimizing for equality times productivity, when the two agents with the highest build-skill (Agents 1, 2) are (not) in the top half, taxes in the $204$ to $510$ bracket are lower (higher) than those in the $0$ to $84$ range.

Owing to the complexity of these environments, it is not possible to provide an intuitive explanation of these AI tax schedules.
%%%%%%%%%%%%%%%%
Nevertheless, it is not surprising that differences between scenarios reflect in the optimal tax rates, as the various combinations of skill and resource access promote difference economic forces and resulting equilibria.
%%%%%%%%%%%%%%%%
Such is demonstrated even in the range of free market social outcomes across these scenarios (\ExtendedDataString{}Figure \ref{ext-data-fig:gtb-quantative-results}d,g).
%%%%%%%%%%%%%%%%
Considering that the AI tax schedules maximize social welfare within their respective scenarios, we view their scenario-specific idiosyncrasies as evidence of the adaptability of the AI Economist framework.

%%%%%%%%%%%%%%%%
%%%%%%%%%%%%%%%%
%%%%%%%%%%%%%%%%
\begin{figure}[t!]
    \centering
    \includegraphics[width=\linewidth]{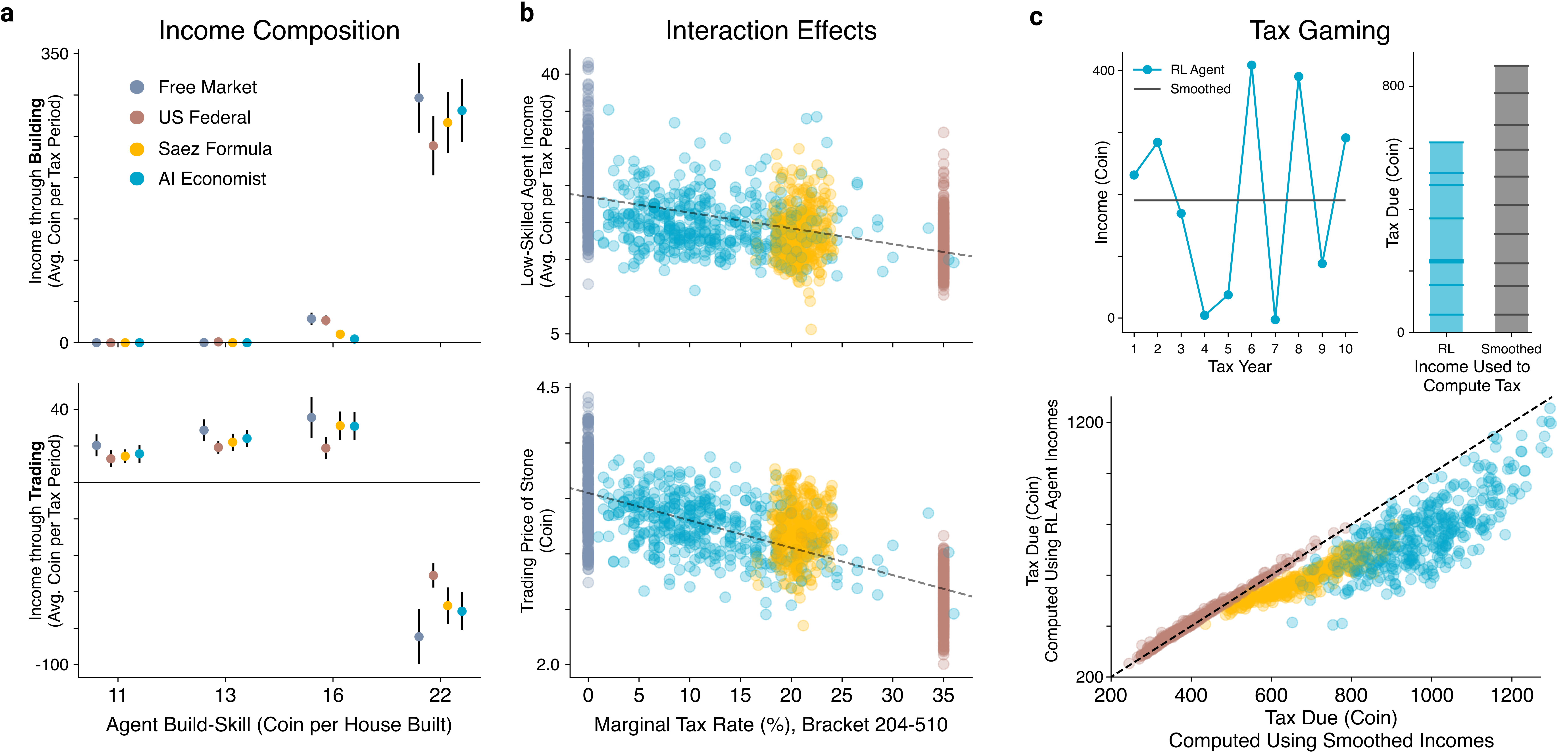}
    \caption{
        \textbf{Specialization, interactions, and tax gaming in the 4-agent Open-Quadrant \GatherTradeBuild{} environment.}
        %%%%%%%%%%%%%%%%
        \textbf{\figfivea,} Average net income from building (\figfivea, top) and trading (\figfivea, bottom) of each agent. Negative values denote net expenditure.
        %%%%%%%%%%%%%%%%
        \textbf{\figfiveb,} The income of the two lowest build-skill agents (\figfiveb, top) and average trading price (\figfiveb, bottom) decrease as the  tax rate in the higher $204$-$510$ tax bracket increases, even though the agents' incomes are below the cutoff for this bracket.
        %%%%%%%%%%%%%%%%
        Hence, the trading behavior of high-skilled agents affects the income of the low-skilled agents.
        %%%%%%%%%%%%%%%%
        The standard definition of elasticity does not capture this interaction effect.
        %%%%%%%%%%%%%%%%
        \textbf{\figfivec,} RL agents learn to strategize across each of the 10 tax years, lowering their total payable tax compared to a smoothed strategy that earns the same, average income in each year:
        %%%%%%%%%%%%%%%%
        the top panels illustrate this for a single episode; the bottom panel shows the saving relative to a smoothed income across all episodes used in the analysis. We do not observe this tax gaming under the progressive US Federal tax schedule.
    }
    \label{fig:economic-interactions-and-income-elasticity}
\end{figure}

\hypertarget{policy-design-beyond-independence-assumptions}{
    \section{
    Policy Design Beyond Independence Assumptions
    }\label{policy-design-beyond-independence-assumptions}
}
%%%%%%%%%%%%%%%%
%%%%%%%%%%%%%%%%
Micro-founded AI-driven simulations such as \GatherTradeBuild{} enable optimal tax policy design in multi-step economies with coupled agent behaviors and interactions, through two-level RL.
%%%%%%%%%%%%%%%%
In contrast, analytical solutions are not available for these kinds of environments: traditional methods fail to account for interactions and thus only achieve suboptimal social welfare.

To illustrate the effect of interactions, Figure \ref{fig:economic-interactions-and-income-elasticity}a-b shows that the income of the two agents with the lowest build-skill depends on the second-to-highest bracket tax rate, even though this income bracket only directly applies to the agent with the highest build-skill.
%%%%%%%%%%%%%%%%
As this tax rate increases, the agent with the highest build-skill buys fewer resources.
%%%%%%%%%%%%%%%%
In turn, the average resource price as well as the trade volume decreases, reducing the incomes of the low build-skill agents.
%%%%%%%%%%%%%%%%
%%%%%%%%%%%%%%%%
Hence, a behavioral change of one agent can change the optimal policy of another agent.
%%%%%%%%%%%%%%%%

However, the Saez analysis uses assumptions and a standard definition of elasticity that fail to account for interactions that arise in multi-step (real-world) economies, these interactions arising through trading for example.
%%%%%%%%%%%%%%%%
The Saez analysis assumes that behavioral changes of agents are independent and do not affect each other.
%%%%%%%%%%%%%%%%
This limitation results in suboptimal policy and lost social welfare under the Saez tax, when applied to the \GatherTradeBuild{} environment.

To illustrate this, for the four agent, \textit{Open Quadrant} scenario, a typical regression of observed taxes paid and reported incomes would estimate elasticity at around $0.87$, see Methods for details.
%%%%%%%%%%%%%%%%
However, by evaluating the Saez tax over a wide range of elasticity values, we find that an assumed elasticity of around $3$ optimizes social welfare when used in Saez's framework.
%%%%%%%%%%%%%%%%
This mismatch between offline estimates and imputed optimal values for agent elasticity is in significant part due to interactions between agents.

%% file: src/v2/discussion.tex
\hypertarget{discussion-200-words}{
    \section{
        Discussion
    }\label{discussion-200-words}
}
The AI Economist demonstrates for the first time that economic policy design using RL, together with principled economic simulation, is sound, viable, flexible, and effective.
It suggests an exciting research agenda: using AI to enable a new approach to economic design.
The AI Economist framework can be used to study different policy goals and constraints, and, as AI-driven simulations grow in sophistication, may help to address the modern economic divide.
In particular, AI-driven simulations enable economic policies to be tested in more realistic environments than those available to analytical methods, and show promise in validating assumptions in policy proposals and evaluating ideas coming from economic theory.

However, these results are a first step and are not ready to be implemented as real-world policy. Future research should scale up AI-driven simulations and calibrate them to real-world data, along with learning AI policies that are explainable and robust to simulation-to-reality gaps.
Also, designing simulations to incorporate different societal values and be representative of different parts of society will be an important direction for future work.

AI-driven policy design could democratize policymaking, for instance, through easily accessible open-source code releases that enable a broad multidisciplinary audience to inspect, debate, and build future policymaking frameworks.
As such, we hope the potential of AI-driven policy design will motivate building fair and inclusive data, computation, and governance structures that ultimately improve the social good.

%% file: src/v2/ethics.tex
\hypertarget{Ethics-Discussion}{
    \section{
        Ethics
    }\label{Ethics-Discussion}
}
While the current version of the AI Economist provides only a limited representation of the real world, we recognize that it could be possible to manipulate future, large-scale iterations of the AI Economist to increase inequality and hide this action behind the results of an AI system.

Furthermore, either out of ignorance or malice, bad training data may result in biased policy recommendations, particularly in cases where users will train the tool using their own data. For instance, the under-representation of communities and segments of the work-force in training data might lead to bias in AI-driven tax models. This work also opens up the possibility of using richer, observational data to set individual taxation, an area where we anticipate a strong need for robust debate.

Economic simulation enables studying a wide range of economic incentives and their consequences, including models of stakeholder capitalism.
However, the simulation used in this work is not an actual tool that can be currently used with malintent to reconfigure tax policy.
We encourage anyone utilizing the AI Economist to publish a model card and data sheet that describes the ethical considerations of trained AI-driven tax models to increase transparency, and by extension, trust, in the system.
Furthermore, we believe any future application or policy built on economic simulations should be built on inspectable code and subject to full transparency.

In order to responsibly publish this research, we have taken the following measures:

\begin{itemize}
\item To ensure accountability on our part, we have consulted academic experts on safe release of code and ensured we are in compliance with their guidance. We shared the paper and an assessment of the ethical risks, mitigation strategies, and assessment of safety to publish with the following external reviewers:
Dr. Simon Chesterman, Provost’s Chair and Dean of the National University of Singapore Faculty of Law, and Lofred Madzou, AI Project Lead at the World Economic Forum's Center for the Fourth Industrial Revolution.
None of the reviewers identified additional ethical concerns or mitigation strategies that should be employed. All affirmed that the research is safe to publish.
\item To increase transparency, we are also publishing a summary of this work as a blog post, thereby allowing robust debate and broad multidisciplinary discussion of our work.
\item To further promote transparency, we will release an open-source version of our environment and sample training code for the simulation.
This does not prevent future misuse, but we believe, at the current level of fidelity, transparency is key to promote grounded discussion and future research.
\end{itemize}

With these mitigation strategies and other considerations in place, we believe this research is safe to publish.
Furthermore, this research was not conducted with any corporate or commercial applications in mind.

%% file: src/v2/methods.tex
\begin{figure}[t!]
    \centering
    \includegraphics[width=\linewidth]{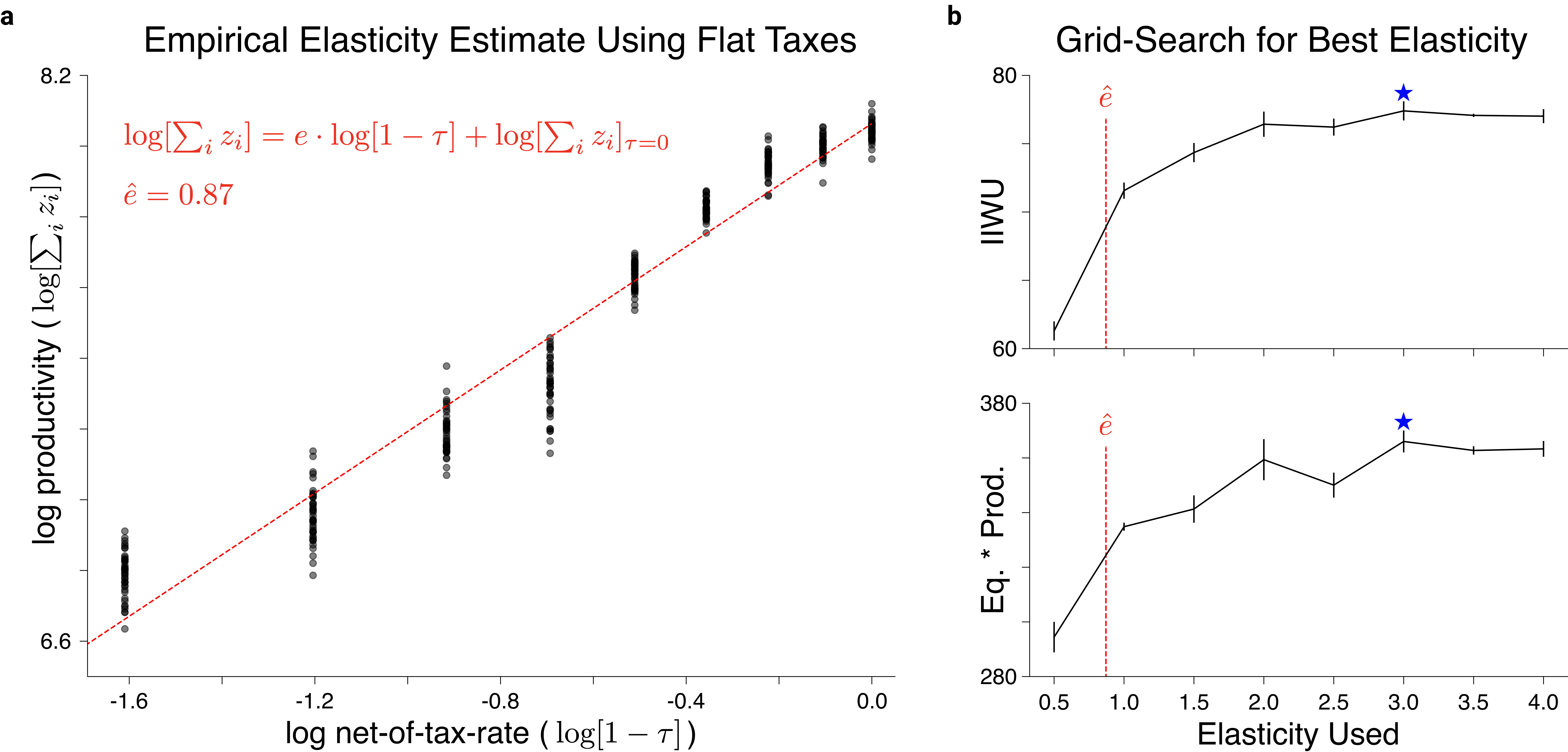}
    \caption{
        \textbf{Estimating elasticity for the Saez tax in the 4-agent Open-Quadrant scenario.}
        \textbf{a,} Regression on income and marginal tax-rate data yields elasticity estimates $\hat{\elas}$ of approximately 0.87 (slope of the red dotted line).
        The net-of-tax-rate ($1 - \tau$) is the fraction of income agents retain after paying taxes.
        Productivity ($\sum_i z_i$) is the total pre-tax income earned by the agents.
        Each dot represents a $(\sum_i \income_i, \mtaxrate)$ pair observed from a sweep over flat tax rates (see Methods).
        \textbf{b,} Social welfare with agents trained to convergence under the Saez tax, using a grid-search over elasticity parameters. Social welfare is highest under the Saez tax when the used elasticity parameter is approximately 3 (blue star), for both the inverse-income-weighted-utility objective (top) and the equality-times-productivity objective (bottom).
        Error bars denote standard error across the 3 random seeds used for each elasticity value.
    }
    \label{ext-data-fig:income-elasticity}
\end{figure}

\hypertarget{methods}{
    \section{
        Methods
    }\label{methods}
}
\paragraph{Dataset Use and Availability.}
No independent, third-party datasets were used in this work. All results were obtained through the use of simulation.
The data and code used to visualize the results is available for all Figures, specifically:
\begin{itemize}
    \item Figure \ref{fig:emergent-phenomena}
    \item Figure \ref{fig:quantitative-results}
    \item Figure \ref{fig:ai-tax-policy-wealth-transfer-and-strategies}
    \item Figure \ref{fig:economic-interactions-and-income-elasticity}
    \item \ExtendedDataString{}Figure \ref{ext-data-fig:gtb-quantative-results}
    \item \ExtendedDataString{}Figure \ref{ext-data-fig:income-elasticity}
\end{itemize}
\paragraph{Code Availability.}
All code for the economic simulations, reinforcement learning algorithms, and analysis are available upon request from the corresponding author.
\paragraph{One-Step Economy.}
We trained the AI Economist in a stylized, one-step economy with $\numberofagents = 100$ agents, indexed by $i$, that each choose how many hours of labor $\labor_i$ to perform. Each agent $i$ has a \emph{skill level} $\skill_i$, which is a private value that represents its hourly wage. Based on labor, each agent $i$ earns a pre-tax income $\income_i = \labor_i \cdot \skill_i$. Each agent $i$ also pays income tax $\tax(\income_i)$ which is evenly redistributed back to the agents. As such, the post-tax income is defined as $\posttax{\income}_i = \income_i - \tax(\income_i) + \frac{1}{\numberofagents}\sum_{j=1}^\numberofagents \tax(\income_j)$.
As a result, each agent $i$ experiences a \emph{utility} $\util(\posttax{\income}_i, \labor_i) = \posttax{\income}_i - \laborcoeff \cdot \labor_i^{\laborexponent}$, which increases linearly with post-tax income $\posttax{\income}_i$ and decreases exponentially with labor $\labor_i$, with exponent $\laborexponent > 0$ and constant $\laborcoeff > 0$ (for exact values used, see \ExtendedDataString{}Table \ref{ext-data-tab:one-step-economy-hyperparameters}).
%%%%%%%%%%%%%%
%%%%%%%%%%%%%%
%%%%%%%%%%%%%%
\begin{table}[ht]
    \begin{small}
        \begin{center}
            {\sffamily % <===========================================================
            \begin{tabular}[c]{lll}
                \hline
                Parameter & & Value \\
                \hline
                Number of agents & $\numberofagents$ & 100 \\
                Minimum skill value & & 1.24 \\
                Maximum skill value & & 159.1 \\
                Maximum labor choice & & 100 \\
                Labor disutility coefficient & $\laborcoeff$ & 0.0005 \\
                Labor disutility exponent & $\laborexponent$ & 3.5 \\
                Min bracket rate & & 0\% \\
                Max bracket rate & & 100\% \\
                Rate discretization (AI Economist) & & 5\% \\
                \hline
            \end{tabular}
            } % <====================================================================
        \end{center}
        \caption{Hyperparameters for the One-Step Economy environment.}
        \label{ext-data-tab:one-step-economy-hyperparameters}
    \end{small}
\end{table}
%%%%%%%%%%%%%%
%%%%%%%%%%%%%%
%%%%%%%%%%%%%%
\begin{table}[ht]
    \begin{small}
        \begin{center}
            {\sffamily % <===========================================================
            \begin{tabular}[c]{lll}
                \hline
                Parameter & & Value \\
                \hline
                Episode length & $\eplen$ & 1000 \\
                World height & $\height$ & 25 \\
                World width & $\width$ & 25 \\
                Resource respawn probability & & 0.01 \\
                Max resource health & & 1 \\
                Starting agent coin & $\money_{i, 0} $ & 0 \\
                Iso-elastic utility exponent & $\isoeta$ & 0.23 \\
                Move labor & & 0.21 \\
                Gather labor & & 0.21 \\
                Trade labor & & 0.05 \\
                Build labor & & 2.1 \\
                Minimum build payout & & 10 \\
                Build payment max skill multiplier & & 3 \\
                Max bid/ask price & & 10 \\
                Max bid/ask order duration & & 50 \\
                Max number of open orders per resource & & 5 \\
                Tax period duration & $\taxperiodlen$ & 100 \\
                Min bracket rate & & 0\% \\
                Max bracket rate & & 100\% \\
                Rate discretization (AI Economist) & & 5\% \\
                \hline
            \end{tabular}
            } % <====================================================================
        \end{center}
        \caption{Hyperparameters for the \GatherTradeBuild{} environment.}
        \label{ext-data-tab:env-hyperparameters}
    \end{small}
\end{table}
%%%%%%%%%%%%%%
%%%%%%%%%%%%%%
%%%%%%%%%%%%%%
\paragraph{\GatherTradeBuild{} Simulation.}
\emph{\GatherTradeBuild{}} simulates a multi-step trading economy in a two-dimensio\-nal grid-world. \ExtendedDataString{}Table \ref{ext-data-tab:env-hyperparameters} provides details regarding the simulation hyperparameters.
Agents can gather resources, earn coins by using the resources of stone and wood to build houses, and trade with other agents to exchange resources for coins.
Agents start at different initial locations in the world and are parameterized by different skill levels (described below).
Simulations are run in episodes of 1000 timesteps, which is subdivided into 10 tax periods, each lasting 100 timesteps.

The state of the world is represented as an $\height \times \width \times \channels$ tensor, where $\height$ and $\width$ are the size of the world and $\channels$ is the number of unique entities that may occupy a cell, and the value of a given element indicates which entity is occupying the associated location.

The action space of the agents includes 4 movement actions: up, down, left, and right. Agents are restricted from moving onto cells that are occupied by another agent, a water tile, or another agent's house.

Stone and wood stochastically spawn on special resource regeneration cells. Agents can gather these resources by moving to populated resource cells.
After harvesting, resource cells remain empty until new resources spawn. By default, agents collect 1 resource unit, with the possibility of a bonus unit also being collected, the probability of which is determined by the agent's \textit{gather-skill}.
Resources and coins are accounted for in each agent's \emph{endowment} $\endow$, which represents how many coins, stone, and wood each agent owns.

\begin{figure}[t!]
    \centering
    \includegraphics[width=.75\linewidth]{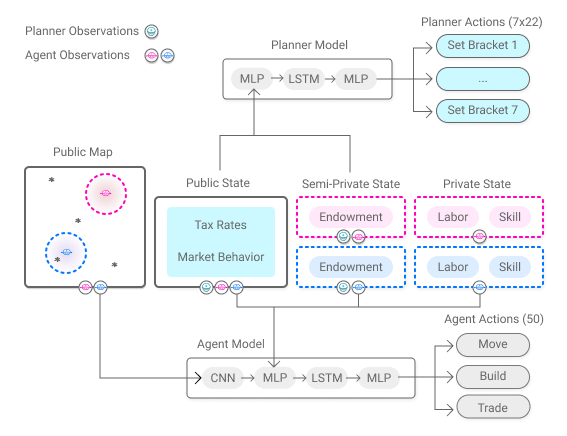}
    \caption{
        \textbf{Observation and action spaces for economic agents and the social planner.}
        The agents and the planner observe different subsets of the world state.
        Agents observe their spatial neighborhood, market prices, tax rates, inventories, and skill level.
        Agents can decide to move (and therefore gather if moving onto a resource), buy, sell, or build.
        There are 50 unique actions available to the agents.
        The planner observes market prices, tax rates, and agent inventories.
        The planner decides how to set tax rates, choosing one of 22 settings for each of the 7 tax brackets.
    }
    \label{ext-data-fig:observation-action}
\end{figure}

Agent observations include the state of their own endowment (wood, stone, and coin), their own skill levels, and a view of the world state tensor within an egocentric spatial window (see \ExtendedDataString{}Figure \ref{ext-data-fig:observation-action}).

Our experiments use a world of size 25-by-25 (40-by-40) for four agent (ten agent) environments, where agent spatial observations have size 11-by-11 and are padded as needed when the observation window extends beyond the world grid.

The planner observations include each agent's endowment but not skill levels (see \ExtendedDataString{}Figure \ref{ext-data-fig:observation-action}). We do not include the spatial state in the planner's observations (in pilot experiments, we observed that this choice did not affect performance).
%%%%%%%%%%%%%%%%%%
%%%%%%%%%%%%%%%%%%
%%%%%%%%%%%%%%%%%%
%%%%%%%%%%%%%%%%%%
%%%%%%%%%%%%%%%%%%
%%%%%%%%%%%%%%%%%%
%
\paragraph{Trading.}
Agents can buy and sell resources from one another through a \textit{continuous double auction}. Agents can submit \emph{asks} (the number of coins they are willing to accept) or \emph{bids} (how much they are willing to pay) in exchange for one unit of wood or stone.

The action space of the agents includes 44 actions for trading, representing the combination of 11 price levels ($0, \ldots, 10$ coin), 2 directions (bids and asks), and 2 resources (wood and stone). Each trade action maps to a single order (i.e.~bid 3 coins for 1 wood, ask for 5 coins in exchange for 1 stone, etc.). Once an order is submitted, it remains open until either it is matched (in which case a trade occurs) or it expires (after 50 timesteps). Agents are restricted from having more than 5 open orders for each resource, and are restricted from placing orders that they cannot complete (they cannot bid with more coins than they possess and cannot submit asks for resources that they do not have).

A bid/ask pair forms a valid trade if they are for the same resource and the bid price matches or exceeds the ask price. When a new order is received it is compared against complementary orders to identify potential valid trades. When a single bid (ask) could be paired with multiple existing asks (bids), priority is given to the ask (bid) with the lowest (highest) price; in the event of ties, priority then is given to the earliest order and then at random. Once a match is identified, the trade is executed using the price of whichever order was placed first.

For example, if the market receives a new bid that offers 8 coins for 1 stone and the market has two open asks offering 1 stone for 3 coins and 1 stone for 7 coins, received in that order, the market would pair the bid with the first ask and a trade would be executed for 1 stone at a price of 3 coins.
The bidder loses 3 coins and gains 1 stone; the asker loses 1 stone and gains 3 coins. Once a bid and ask are paired and the trade is executed, both orders are removed.

The state of the market is captured by the number of outstanding bids and asks at each price level for each resource. Agents observe these counts both for their own bids/asks as well as the cumulative bids/asks of other agents. The planner observes the cumulative bids/asks of all agents. In addition, both the agents and the planner observe historical information from the market: the average trading price for each resource, as well as the number of trades at each price level.
\paragraph{Building.}
Agents can choose to spend one unit of wood and one unit of stone to build a house, and this places a house tile at the agent's current location and earns the agent some number of coins. Agents are restricted from building on source cells as well as locations where a house already exists. The number of coins earned per house is identical to an agent's \emph{build-skill}, a numeric value between 10 and 30. As such, agents can earn between 10 and 30 coins per house built. Skill is heterogeneous across agents and does not change during an episode. Each agent's action space includes 1 action for building.
\paragraph{Labor.}
Over the course of an episode of 1000 timesteps, agents accumulate labor cost, which reflects the amount of effort associated with their actions. Each type of action (moving, gathering, trading, and building) is associated with a specific labor cost. All agents experience the same labor costs.
\paragraph{Taxation Mechanism.}
Taxation is implemented using income brackets and bracket tax rates. All taxation is anonymous: tax rates and brackets do not depend on the identity of taxpayers. The payable tax for income $\income$ is computed as follows:
\eq{\label{supp-eq:income-tax}
    \incometax(\income) = \sum_{j=1}^{\numbrackets} \mtaxrate_j \cdot \brck{  \brck{ \bracketcutoff_{j+1} - \bracketcutoff_j } \bm{1}[ \income > \bracketcutoff_{j+1} ]  + \brck{ \income - \bracketcutoff_j } \bm{1}[ \bracketcutoff_j < \income \leq \bracketcutoff_{j+1} ]  },
}
where $\numbrackets$ is the number of brackets, and the $\mtaxrate_j$ and $\bracketcutoff_j$ are marginal tax rates and income boundaries of the brackets, respectively.

Each simulation episode has 10 tax years. On the first time step of each tax year, marginal tax rates are set that will be used to collect taxes when the tax year ends.
For baseline models, tax rates are set either formulaically or fixed. For taxes controlled by a deep neural network, the action space of the planner is divided into seven action subspaces, one for each tax bracket: $\brck{0, 0.05, 0.10, \ldots, 1.0}^7$. Each subspace denotes the set of discretized marginal tax rates available to the planner. Discretization of tax rates only applies to deep learning networks, enabling standard techniques for RL with discrete actions.

Each agent observes the current tax rates, indicators of the temporal progress of the current tax year, and the set of sorted and anonymized incomes the agents reported in the previous tax year.
In addition to this global tax information, each agent also observes the marginal rate at the level of income it has earned within the current tax year so far.
The planner also observes this global tax information, as well as the non-anonymized incomes and marginal tax rate (at these incomes) of each agent in the previous tax year.
\paragraph{Redistribution Mechanism.}
An agent's pretax income $\income_i$ for a given tax year is defined simply as the change in its coin endowment $\money_i$ since the start of the year.
Accordingly, taxes are collected at the end of each tax year by subtracting $\incometax(\income_i)$ from $\money_i$.

Taxes are used to redistribute wealth: the total tax revenue is evenly redistributed back to the agents.
In total, at the end of each tax year, the coin endowment for agent $i$ changes according to $\Delta \money_i = -\incometax(\income_i) + \frac{1}{\numberofagents}\sum_j^\numberofagents\incometax(\income_j)$, where $\numberofagents$ is the number of agents.
Through this mechanism, agents may gain coin when they receive more through redistribution than they pay in taxes.
\paragraph{\GatherTradeBuild{} Scenarios.}
We considered two spatial layouts: \emph{Open-Quadrant} and \emph{Split-World}, see \ExtendedDataString{}Figure \ref{ext-data-fig:gtb-quantative-results}.

Open-Quadrant features four regions delineated by impassable water with passageways connecting each quadrant. Quadrants contain different combinations of resources: both stone and wood, only stone, only wood, or nothing. Agents can freely access all quadrants, if not blocked by objects or other agents.

Split-World features two disconnected regions: the top contains stone and wood, while the bottom only has stone. Water tiles prevent agents from moving from one region to the other.

All scenarios use a fixed set of \textit{build-skills} based on a clipped Pareto distribution (sampled skills are clipped to the maximum skill value) and determine each agent's starting location based on its assigned build-skill.
The Open-Quadrant scenario assigns agents to a particular corner of the map, with similarly skilled agents being placed in the same starting quadrant.
(Agents in the lowest build-skill quartile start in the wood quadrant; those in the second quartile start in the stone quadrant; those in the third quartile start in the quadrant with both resources; and agents in the highest build-skill quartile start in the empty quadrant.)
The Split-World scenario allows control over which agents have access to both wood and stone versus access to only stone.
We consider 4 Split-World variations, each with ten agents.
Each variation gives stone and wood access to a specific subset of the ten agents, as determined by their build-skill rank.
For example: \textit{Split-World-1,2,3} places the 3 highest-skilled agents in the top, \textit{Split-World-8,9,10} places the 3 lowest-skilled agents in the top, and \textit{Split-World-5,6} places the 2 middle-skilled agents in the top.

\paragraph{Agent Utility.}
Following optimal taxation theory, agent utilities depend positively on accumulated coin $\money_{i, t}$, which only depends on post-tax income $\posttax{\income} = \income - \tax(\income)$.
In contrast, the utility for agent $i$ depends negatively on accumulated labor $\totallabor_{i,t} = \sum_{k=0}^t \labor_{i, k}$ at timestep $t$. The utility for an agent $i$ is:
\eq{\label{supp-eq:crra-utility}
    \util_{i,t} = \frac{\money_{i, t}^{1 - \isoeta} - 1}{1 - \isoeta} - \totallabor_{i, t}.
}
Agents learn behaviors that maximize their expected total discounted utility for an episode.
We found that build-skill is a significant determinant of behavior; agents' gather-skill empirically does not affect optimal behavior in our settings.

All of our experiments use a fixed set of build-skills, which, along with labor costs, are roughly calibrated so that (1) agents need to be strategic in how they choose to earn income, and (2) the shape of the resulting income distribution roughly matches that of the 2018 US economy with trained optimal agent behaviors.
\paragraph{Social Planner.}
The simulation environment includes a \emph{social planner} who uses tax policy and lump-sum redistribution to influence social outcomes. Each episode is divided into 10 tax years. At the start of each tax year, the planner chooses a tax schedule $\tax(\income)$ that determines the amount of taxes each agent will owe as a function of its income $\income$ earned during the tax year and redistributes tax revenue.

We compare four kinds of planners:
(1) \emph{Free Market}: a fixed-rate planner where all tax rates are 0\%;
(2) \emph{US Federal}: a fixed-rate planner where bracketed marginal tax rates follow a progressive scheme adapted from the 2018 US federal single-filer income tax schedule;
(3) \emph{Saez tax}: an adaptive planner that computes theoretically optimal marginal rates using the empirical income distribution and elasticity of income with respect to taxation;
and (4) \emph{AI Economist}: a deep neural network, adaptive planner that maps a set of planner observations to bracketed marginal tax rates, which is trained via reinforcement learning (RL) to maximize social welfare.
\paragraph{Two-level Deep Reinforcement Learning.}
RL provides a flexible way to simultaneously optimize and model the behavioral effects of tax policies. We instantiate RL at two levels, that is, for two types of actors: training agent behavioral policy models and a taxation policy model for the social planner.

We train each actor's behavioral policy using deep reinforcement learning, which learns the weights $\policyweight_i$ of a neural network $\policy(\ac_{i,t} | \ob_{i,t}; \policyweight_i)$ that maps an actor's observations to actions.
Network weights are trained to maximize the expected total discounted reward of the output actions.

Specifically, for an agent $i$ using a behavioral policy $\policy_i\brck{\ac_{t}|\ob_{t}; \policyweight_i}$, the RL training objective is (omitting the tax policy $\plannerpolicy$):
\eq{
\max_{\policy_i}\E_{
    \ac_1\sim\policy_1, \ldots, \ac_\numberofagents\sim\policy_\numberofagents, s'\sim\trans
}\brcksq{\sum_{t=0}^\eplen \df^t \rew_t},
}
where $\st'$ is the next state and $\trans$ denotes the dynamics of the environment. The objective for the planner policy $\plannerpolicy$ is similar.
Standard model-free policy gradient methods update the policy weights $\policyweight_i$ using
\eq{\label{methods-eq:policy-gradient}
\Delta\policyweight_i \propto \E_{
    \ac_1\sim\policy_1, \ldots, \ac_\numberofagents\sim\policy_\numberofagents, s'\sim\trans
}\brcksq{\sum_{t=0}^\eplen \df^t \rew_t \nabla_{\policyweight_i} \log\policy_i\brck{\ac_{i,t}|\ob_{i,t}; \policyweight_i}}.
}
In our work, we use \emph{proximal policy gradients} (PPO)~\autocite{schulman2017proximal}, an extension of Formula \ref{methods-eq:policy-gradient} to train all actors (both agents and planner).

To improve learning efficiency, we train a single agent policy network $\policy(\ac_{i,t} | \ob_{i,t}; \policyweight)$ whose weights are shared by all agents, that is, $\policyweight_i = \policyweight$. This network is still able to embed diverse, agent-specific behaviors by conditioning on agent-specific observations.

At each timestep $t$, each agent observes: its nearby spatial surroundings; its current endowment (stone, wood, and coin); private characteristics, such as its building skill; the state of the markets for trading resources; and a description of the current tax rates. These observations form the inputs to the policy network, which uses a combination of convolutional, fully connected, and recurrent layers to represent spatial, non-spatial, and historical information, respectively. For recurrent components, each agent maintains its own hidden state.
This is visualized in \ExtendedDataString{}Figure \ref{ext-data-fig:observation-action}.
For the detailed model architecture and training hyperparameters, see \ExtendedDataString{}Tables \ref{ext-data-tab:training-hyperparameters} and \ref{ext-data-tab:policy-network-hyperparameters}.
%%%%%%%%%%%%
%%%%%%%%%%%%
%%%%%%%%%%%%
\begin{table}[ht]
    \begin{small}
        \begin{center}
            {\sffamily % <===========================================================
            \begin{tabular}[c]{lll}
                \hline
                Parameter & & Value \\
                \hline
                Number of parallel environment replicas & & 30 \\
                Sampling horizon (steps per replica) & $\samplinghorizon$ & 200 \\
                Agent SGD minibatch size (\# agents = 4) & & 600 \\
                Agent SGD minibatch size (\# agents = 10) & & 1500 \\
                Planner SGD minibatch size & & 1500 \\
                SGD sequence length & & 25 \\
                Policy updates per horizon (agent) & & 40 \\
                Policy updates per horizon (planner) & & 4 \\
                CPUs & & 15 \\
                Learning rate (agent) & & 0.0003 \\
                Learning rate (planner) & & 0.0001 \\
                Entropy regularization coefficient (agent) & & 0.025 \\
                Entropy regularization coefficient (planner) & & 0.125 \\
                Discount factor & $\df$ & 0.998 \\
                Generalized Advantage Estimation discount parameter & $\lambda$ & 0.98 \\
                Gradient clipping norm threshold & & 10 \\
                Value function loss coefficient & & 0.05 \\
                Phase \textit{one} training duration & & 25M steps \\
                Phase \textit{two} training duration & & 1B steps \\
                Phase \textit{two} initial max $\mtaxrate$ & & 10\% \\
                Phase \textit{two} tax annealing duration & & 27M steps \\
                Phase \textit{two} entropy regularization annealing duration & & 50M steps \\
                \hline
            \end{tabular}
            } % <====================================================================
        \end{center}
        \caption{Hyperparameters for two-level reinforcement learning (RL), which trains multiple agents and a social planner. The base RL algorithm is the proximal policy gradient algorithm~\autocite{schulman2017proximal}.}
        \label{ext-data-tab:training-hyperparameters}
    \end{small}
\end{table}
%%%%%%%%%%%%
%%%%%%%%%%%%
%%%%%%%%%%%%
\begin{table}[ht]
    \begin{small}
        \begin{center}
            {\sffamily % <===========================================================
            \begin{tabular}[c]{lll}
                \hline
                Parameter & & Value \\
                \hline
                Number of convolutional layers & & 2 \\
                Number of fully-connected layers & & 2 \\
                Fully-connected layer dimension (agent) & & 128 \\
                Fully-connected layer dimension (planner) & & 256 \\
                LSTM cell size (agent) & & 128 \\
                LSTM cell size (planner) & & 256 \\
                Agent spatial observation box half-width & & 5 \\
                \hline
            \end{tabular}
            } % <====================================================================
        \end{center}
        \caption{Hyperparameters for the neural networks implementing the agent and planner policy models.}
        \label{ext-data-tab:policy-network-hyperparameters}
    \end{small}
\end{table}
%%%%%%%%%%%%
%%%%%%%%%%%%
%%%%%%%%%%%%
The policy network for the social planner follows a similar construction, but differs somewhat in the information it observes. Specifically, at each timestep, the planner policy observes: the current inventories of each agent; the state of the resource markets; and a description of the current tax rates. The planner cannot directly observe private information such as an agent's skill level.
\paragraph{Training Objectives.}
Rational economic agents train their policy $\policy_i$ to optimize their total discounted utility over time, while experiencing tax rates $\mtaxrate$ set by the planner's policy $\plannerpolicy$. The agent training objective is:
\eq{\label{supp-eq:agentproblem}
    \forall i: \max_{\policy_i} \E_{
        \mtaxrate\sim\plannerpolicy, \ac_i \sim \policy_i, \Ac_{-i} \sim \bm{\policy}_{-i}, \st'\sim \trans}\brcksq{
        \sum_{t = 1}^\eplen \df^t \rew_{i,t} + \util_{i,0}
    },\quad \rew_{i,t} = \util_{i,t} - \util_{i, t-1},
}
where the instantaneous reward $\rew_{i,t}$ is the marginal utility for agent $i$ at timestep $t$, and we use the isoelastic utility $\util_t$ as defined in Equation \ref{supp-eq:crra-utility}. Bold-faced quantities denote vectors, and the subscript ``$-i$'' denotes quantities for all agents except for $i$.

For an agent population with monetary endowments $\vmoney_t = (\money_{1,t}, \ldots, \money_{N,t})$, we define equality $\equality(\vmoney_t)$ as:
\eq{\label{supp-eq:equality}
    \equality(\vmoney_t) = 1 - \fr{N}{N-1}\gini(\vmoney_t), & \quad 0 \leq \equality(\vmoney_t) \leq 1,
}
where the Gini index is defined as
\eq{\label{supp-eq:gini}
    \gini(\vmoney_t) = \fr{\sum_{i=1}^N \sum_{j=1}^N |\money_{i,t} - \money_{j,t}|}{ 2N \sum_{i=1}^N \money_{i,t}},& \quad 0 \leq \gini(\vmoney_t) \leq \fr{N-1}{N}.
}
We also define productivity as the sum of all incomes:
\eq{\label{supp-eq:productivity}
    \productivity(\vmoney_t) = \sum_i \money_{i,t}.
}
Note that we assume the economy is closed: subsidies are always redistributed evenly among agents, no tax money leaves the system. Hence, the sum of pre-tax and post-tax incomes is the same.
The planner trains its policy $\plannerpolicy$ to optimize social welfare:
\eq{\label{supp-eq:plannerproblem}
    \max_{\plannerpolicy}  \E_{  \mtaxrate \sim \plannerpolicy,  \Ac \sim \bm{\policy},  \st' \sim \trans  }  \brcksq{  \sum_{t=1}^H \df^t \rew_{p,t} + \socialwelfare_0}, \quad
    \rew_{p,t} = \socialwelfare_t - \socialwelfare_{t-1}.
}
The \emph{utilitarian} social welfare objective is the family of linear-weighted sums of agent utilities, defined for weights $\socialwelfareweight_i\geq 0$:
\eq{\label{supp-eq:inv_income_weighted_utility}
    \socialwelfare_t = \sum_{i=1}^N \socialwelfareweight_i \cdot \util_{i,t}.
}
We use inverse-income as the weights: $\socialwelfareweight_i \propto \frac{1}{\money_i}$, normalized to sum to 1.
We also adopt an objective that optimizes a trade-off between equality and productivity, defined as the product of equality and productivity:
\eq{\label{supp-eq:eqprodswf}
    \socialwelfare_t = \equality(\vmoney_t) \cdot \productivity(\vmoney_t).
}
As agent incomes $\income_i$ depend on skill and access to resources, the heterogeneity in initial locations and build-skill are the main drivers of both economic inequality and specialization in \GatherTradeBuild{}.
\paragraph{Training Strategies.}
Two-level RL can be unstable, as the planner's actions (setting tax rates) affect agent rewards (marginal utility depending on post-tax income).

We employ three learning curricula and two training phases to stabilize two-level RL. In phase one, agent policies are trained from scratch in a free-market (no-tax) environment for 50 million steps. In phase two, agents continue to learn in the presence of taxes for another 1 billion steps.

The first learning curriculum occurs during phase one: agents use a curriculum in phase one that anneals the utility cost associated with labor. The reason is that many actions cost labor, but few yield income. Hence, if exploring without a curriculum, a suboptimal policy can experience too much labor cost and converge to doing nothing.

The second learning curriculum occurs during phase two: we anneal the maximum marginal tax to prevent planners from setting extremely high taxes during exploration that reduce post-tax income to zero and discourage agents from improving their behaviors.

We also carefully balance entropy regularization, to prevent agent and planner policies from prematurely converging and promote the co-adaption of agent and planner policies. The entropy of the policy $\policy$ for agent $i$, given an observation $\ob_i$, is defined as:
\eq{\label{supp-eq:entropy}
   \texttt{entropy}(\policy) = -\E_{\ac\sim\policy}\brcksq{\log\policy(\ac | \ob_{i} ; \policyweight_i)}.
}
When training the AI Economist planner, we introduce the third learning curriculum by annealing the level of planner policy entropy regularization. Enforcing highly entropic planner policies during the early portion of phase two allows the agents to learn appropriate responses to a wide range of tax levels before the planner is able to optimize its policy.
\paragraph{Training Procedure.}
For training, we use proximal policy gradients (PPO) on mini-batches of experience collected from 30 parallel replicas of the simulation environment. Each environment replica runs for 200 steps during a training iteration.
Hence, for each training iteration, 6,000 transitions are sampled for the planner and $\numberofagents \cdot$ 6,000 transitions are sampled for the agents, where $\numberofagents$ is the number of agents in the scenario, using the latest policy parameters.

The planner policy model is updated using transition mini-batches of size 1500, with one PPO update per minibatch (4 updates per iteration). The agent policy model is updated using transition mini-batches of size 400 (1500) for 4 (10) agent scenarios (40 updates per iteration).
\ExtendedDataString{}Table \ref{ext-data-tab:policy-network-hyperparameters} provides details regarding the training hyperparameters.
\ExtendedDataString{}Algorithm \ref{ext-data-alg:two-level-rl} describes the full training procedure.
\paragraph{Action Spaces and Masks.}
Both agents and planners use discrete action spaces. We use action masking to prevent invalid actions, e.g., when agents cannot move across water, and to implement learning curricula. Masks control which actions can be sampled at a given time by assigning zero probability to restricted actions.

In addition, we include a no-operation action (\texttt{NO-OP}) in each action space. For the planner, each of the 7 action subspaces includes a \texttt{NO-OP} action. The \texttt{NO-OP} action allows agents to idle and the planner to leave a bracket's tax rates unchanged between periods.

Action masks allow the planner to observe every timestep while only acting at the start of each new tax year. After the first timestep of a tax year, action masks enforce that only \texttt{NO-OP} planner actions are sampled.
\paragraph{Saez Tax.}
The Saez tax computes tax rates using an analytical formula~\autocite{saez_using_2001} for a one-step economy with income distribution $\pdf(\income)$ and cumulative distribution $\cdf(\income)$. These rates maximize a weighted average $\sum_i w_i u_i$ of agent utilities, where the weights $w_i$ reflect the redistributive preferences of the planner, and are optimal in an idealized one-step economy. The Saez tax computes marginal rates as:
\eq{\label{supp-eq:saez-tax}
    \tau(\income) = \frac{1 - G(\income)}{1 - G(\income) + a(\income)  \elas(\income)},
}
where $\income$ is pre-tax income, $G(\income)$ is an income-dependent social welfare weight and $a(z)$ is the local Pareto parameter.

Specifically, let $\alpha(\income)$ denote the \emph{marginal average income at income $\income$}, normalized by the fraction of incomes above $\income$, i.e.,
\eq{\label{supp-eq:saez-alpha}
    \alpha(\income) &= \frac{\income \cdot \pdf(\income)}{1-\cdf(\income)}.
}
Let $G(\income)$ denote the \emph{normalized, reverse cumulative Pareto weight} over incomes above a threshold $\income$, i.e.,
\eq{\label{supp-eq:saez-g}
    G(\income) = \frac{1}{1 - \cdf\brck{\income}} \int_{\income'=\income}^{\infty} p(\income') g(\income') \mathrm{d}\income'.
}
where $g(\income)$ is the normalized social marginal welfare weight of an agent earning income $\income$, and $1 - \cdf(\income)$ is the fraction of incomes above income $\income$.
In this way, $G(\income)$ represents how much the social welfare function weights the incomes above $\income$.
Let \emph{elasticity} $\elas(\income)$ denote the \emph{sensitivity of an agent's income to changes in the tax rate when that agent's income is $\income$}, defined as
\eq{\label{supp-eq:elasticity}
     \elas(\income) = \frac{1-\mtaxrate(\income)}{\income} \frac{\mathrm{d}\income }{\mathrm{d}(1-\mtaxrate(\income)) }.
}
Both $G(\income)$ and $a(z)$ can be computed directly from the (empirical) income distribution, but typically $e(\income)$ needs to be estimated (which is challenging).

We set the social welfare weights $w_i \propto \frac{1}{\income_i}$, normalized so the sum over all individuals is 1. This choice encodes a welfare focus on low-income agents.
\paragraph{Empirical Income Distribution.}
To apply the Saez tax, we use rollout data from a temporal window of episodes to estimate the empirical income distribution and compute $G(\income)$ and $a(z)$. We aggregate reported incomes over a look-back window. We maintain a buffer of recent incomes reported by the agents, where each data point in this buffer represents the income reported by a single agent during a single tax year. Each simulation episode includes 10 tax years. As such, a single agent may report incomes in multiple different brackets in a single episode.

To compute $G(\income)$ and $a(\income)$, we first discretize the empirical income distribution and compute $\tau(\income)$ within each of the resulting income bins. To get the average tax rate $\tau$ for each tax bracket, we take the average of the binned rates over the bracket's income interval. Following the Saez analysis~\autocite{saez_using_2001}, when computing the top bracket rate, $G(\income)$ is the total social welfare weight of the incomes in the top bracket, and $a(\income)$ is computed as $\frac{m}{m-\income^+}$ where $m$ is the average income of those in the top bracket, and $\income^+$ is the income cutoff for the top bracket (510 in our implementation, see Figure \ref{fig:ai-tax-policy-wealth-transfer-and-strategies}.
\paragraph{Estimating Elasticity.}
The most substantial obstacle to implementing the Saez tax is correctly identifying the elasticity $\elas(\income)$, defined as in Equation \ref{supp-eq:elasticity}.
Owing to the complexity of the \GatherTradeBuild{} economy and agent learning dynamics, it is challenging to reliably measure local elasticities $\elas(\income)$ as a function of income $\income$. The large variance in empirical incomes caused large variance in the estimated local elasticity, leading to unstable two-level RL.

Therefore, we used a \emph{global} elasticity estimate $\elas$, which assumes that elasticity is the same at all income levels. Empirically, we observe that the elasticity does not vary greatly across income ranges, hence justifying using a global elasticity.

For comparison, we also estimated the elasticity $\elas(\income)$ using classic techniques, which use regression on observed incomes and marginal tax-rates obtained from agents trained under varying fixed flat-tax systems~\autocite{gruber2002elasticity}.
Using a global constant elasticity for all agents, we instantiate this method by regression on $K$ tuples $\brcksq{\brck{\totalincome_k, \mtaxrate_k}}_{k=1}^{K}$ of observed total income $\totalincome = \sum_i \income_i$ and manually fixed flat tax rates $\mtaxrate$ in the simulation. Specifically, we use a linear model:
\eq{\label{supp-eq:elasticitysimpleformula_for_ols}
    \log (\totalincome) = \hat{\elas} \cdot \log (1-\mtaxrate) + \log ( \hat{\totalincome}^0),
}
where $\hat{\totalincome}^0$ is a bias unit.
Using a flat tax rate ensures agents always face the same tax rate during episodes, allowing for more consistent estimates.
To generate data, we sweep over a range of values for $\mtaxrate$, and collect observed total income data $\totalincome$.
This yields an estimate of $\hat{\elas} \sim 1$, which produces suboptimal social welfare. See \ExtendedDataString{}Figure \ref{ext-data-fig:income-elasticity}.

To provide the best possible performance for the Saez framework, we optimize the Saez tax using a grid search over possible $\elas$ values.
For each scenario, we separately conduct experiments involving sweeps over a range of potential values of $\elas$ and select the best-performing one for each social welfare objective to use as a fixed elasticity estimate. This yields optimal elasticity estimate $\elas \sim 3$ in the 4-agent Open-Quadrant scenario, substantially higher than that estimated through regression techniques. See \ExtendedDataString{}Figure \ref{ext-data-fig:income-elasticity}.
%%%%%%%%%%%%%%%%%
%%%%%%%%%%%%%%%%%
%%%%%%%%%%%%%%%%%
%%%%%%%%%%%%%%%%%
%%%%%%%%%%%%%%%%%
%%%%%%%%%%%%%%%%%
\paragraph{Quantification and Statistical Significance.}
All experiments in the Open-Quadrant \GatherTradeBuild{} scenarios were repeated with 10 random seeds; experiments in the Split-World \GatherTradeBuild{} scenarios and the One-Step Economy were repeated with 5 random seeds.

For a given repetition, we compute each performance metric, e.g. equality or social welfare, as its average value over the last 3000 episodes of training (the last 100 episodes for each of the 30 parallel environments).
We report the average and standard error of these metrics across the 5 or 10 random seeds within a particular experiment group (Figure \ref{fig:quantitative-results}, Figure \ref{ext-data-fig:gtb-quantative-results}).
Statistical significance is computed using a two-sample t-test.

In other analyses (Figures \ref{fig:ai-tax-policy-wealth-transfer-and-strategies} and \ref{fig:economic-interactions-and-income-elasticity}), we compute agent-wise statistics, e.g. pre-tax income and wealth transfer, using agent-specific statistics for each of the 10 tax periods in the episode.
We conduct our analyses using the 40 most recent episodes (prior to the end of training, or prior to 250 million training steps where noted) for each repetition.
For these analyses, we report the averages and standard deviations across the 400 associated episodes within each group of experiments.

%% file: src/v2/end-notes.tex
\hypertarget{endnotes}{%
\section{End Notes}\label{endnotes}}
\begin{itemize}
\item \textbf{Acknowledgements.}
    We thank Kathy Baxter for the ethical review. We thank Nikhil Naik, Lofred Madzou, Simon Chesterman, Rob Reich, Mia de Kuijper, Scott Kominers, Gabriel Kriendler, Stefanie Stantcheva, Stefania Albanesi, and Thomas Piketty for valuable discussions.
\item \textbf{Author Contributions.}
    A.T. and S.Z. contributed equally. R.S. and S.Z. conceived and directed the project; S.Z., A.T., and D.P. developed the theoretical framework; A.T., S.S., and S.Z. developed the economic simulator, implemented the reinforcement learning platform, and performed experiments; A.T., S.Z., and D.P. processed and analyzed experiments with AI agents; S.Z., A.T., and D.P. drafted the manuscript; R.S. planned and advised the work, and analyzed all results; All authors discussed the results and commented on the manuscript.
\item Source code for the economic simulation is available at
    \url{https://www.github.com/salesforce/ai-economist}.
\item The authors declare no competing interests.
\item All data needed to evaluate the conclusions in the paper are present in the paper and/or the Supplementary Materials.
\item The data can be provided by Stephan Zheng pending scientific review and a completed material transfer agreement. Requests for the data should be submitted to: \\ stephan.zheng@salesforce.com.
\item The authors acknowledge that they received no funding in support for this research.
\end{itemize}